\documentclass[letterpaper,journal]{IEEEtran}
\usepackage{amsmath,amsfonts}
\usepackage{algorithmic}
\usepackage{algorithm}
\usepackage{array}
\usepackage{textcomp}
\usepackage{stfloats}
\usepackage{url}
\usepackage{verbatim}
\usepackage{graphicx}
\usepackage{xcolor}
\usepackage{caption}
\usepackage{subcaption}
\usepackage{multirow}
\usepackage{tabularx}
\usepackage{enumitem}
\usepackage{float}
\usepackage{booktabs}
\usepackage{microtype}
\usepackage{placeins}
\usepackage{hyperref}
\usepackage[table]{xcolor}
\usepackage{colortbl}
\usepackage{comment}
\newcommand{\mycc}[1]{\cellcolor{yellow!30}#1}
\newcolumntype{Y}{>{\raggedright\arraybackslash}X}
\newcommand{\paperplotdir}{figures/paper2}
\newcommand{\cropplotdir}{figures/cropping_examples}
\newcommand{\ufprplotdir}{figures/ufpr_examples}

\begin{document}

\title{Efficient Periocular Recognition through Model Scaling and Cross-Spectral VIS–NIR Verification}
\author{
    Fabian Stockhard, Juan~E.~Tapia,~\IEEEmembership{Senior Member,~IEEE,}
    and~Christoph~Busch,~\IEEEmembership{Fellow~Member,~IEEE}%
\thanks{Fabian Stockhard, Juan Tapia and Christoph Busch are with the da/sec-Biometrics and Internet Security Research Group, Hochschule Darmstadt, Germany, e-mail: \{juan.tapia-farias, christoph.busch\}@h-da.de.}%
\thanks{Manuscript received Month DD, YYYY; revised Month DD, YYYY.}}

\maketitle

\begin{abstract}

Periocular recognition can provide a computationally efficient alternative to full-face verification, particularly when facial regions are occluded or image resolution is limited. This work presents \textit{AdaEye}, a systematic development of input and backbone scaling for periocular verification under visible-spectrum (VIS) and cross-spectral conditions (VIS-NIR). Starting from an AdaFace-style face-recognition framework, we compare full-face models, masked-face inputs, and dedicated $80\times40$ periocular models while varying backbone width and embedding dimensionality. Experiments cover four VIS–VIS and VIS–NIR verification protocols and evaluate biometric accuracy together with computational cost, inference throughput, training memory, template size, and training stability. Direct periocular processing substantially improves deployment efficiency, reducing recognition cost from $6.30$ to $1.64$ GMACs per image while increasing measured throughput by approximately three times. The results further demonstrate that its effectiveness is strongly linked to the evaluation protocol: direct periocular models deliver improved performance on native UFPR periocular images, while face-based protocols favor retaining more input information. Moderate backbone-width reductions maintain comparable mean verification performance on \textit{CASIA} and \textit{ND-NIVL}, indicating that these settings can achieve meaningful computational savings without sacrificing accuracy. Embedding compression further reduces template-storage requirements, with a primarily practical impact on deployment. Multi-seed experiments also identify configurations requiring careful numerical stabilization, particularly when embedding and backbone reductions are combined. Taken together, these findings provide a reliable basis for selecting compact periocular architectures according to the target protocol and deployment constraints, balancing verification accuracy, computational efficiency, storage requirements, and robustness.
\end{abstract}


\section{Introduction}
\IEEEPARstart{B}{iometric} recognition systems are increasingly required to operate under real-world constraints: partial visibility, limited computational resources, mobile or edge deployment, and heterogeneous acquisition conditions \cite{Anjith2026}. While full-face recognition achieves high accuracy in controlled settings, it is not always feasible. In scenarios involving occlusion (e.g., masks, make-up, glasses), suboptimal camera framing, privacy-preserving data minimization, or dynamic biometric capture (e.g., on-the-move), the periocular region often becomes the most reliable and delivers complementary accessible biometric cues.

The periocular region, encompassing the eyes, eyelids, eyebrows, and surrounding facial skin, offers a compact yet rich biometric trait, bridging the gap between full-face and iris-based recognition. It retains discriminative identity information even when the lower face is obscured, while requiring significantly less acquisition precision than high-resolution iris recognition. In this work, we treat periocular recognition as a reduced-input face recognition problem, rather than a dedicated iris recognition task. Although the iris may be visible within the periocular crop and complementary facial cues are present, we do not assume the availability of iris-recognition-quality texture. Our focus is on robust, efficient recognition using aligned face for periocular crops, particularly in VIS–NIR cross-spectral scenarios, where the input is constrained but sufficient for identity verification.

Based on these limitations, we address two central questions:
\begin{enumerate}
\item How do input-region reduction and subsequent scaling of backbone depth, channel width, embedding dimensionality, and final spatial representation affect the accuracy-efficiency trade-off in periocular recognition?
\item How do these trade-offs vary across evaluation protocols, operating points, and training seeds, including the reliability of training under a fixed protocol?
\end{enumerate}

The general-purpose EfficientNet network~\cite{pmlr-v97-tan19a} has demonstrated that input resolution, network depth and channel width must be jointly optimized to achieve optimal accuracy-efficiency trade-offs across several tasks. Following this principle, we adopt a systematic framework for periocular recognition.

Instead of introducing a new backbone architecture, we develop \textit{AdaEye} as a dedicated periocular-recognition model. We adapt, modify, and train the \textit{AdaFace} network from scratch to investigate the effects of key architectural and representational choices. This controlled design enables us to isolate the contribution of each dimension and identify efficient configurations tailored specifically to hardware restricted periocular biometrics. \textit{All the models will be available in GitHub {upon acceptance}}.

The contribution of this work is a controlled accuracy-efficiency study for periocular recognition:

\begin{itemize}[leftmargin=*]
    \item A unified \textit{AdaEye} framework trained from scratch enables fair comparison across full-face, masked periocular, and direct $80\times40$ periocular inputs, demonstrating adaptability to efficient, reduced-input periocular recognition in VIS/NIR.
    
    \item Comprehensive evaluation for biometric performance~(FNMR@FMR) and key computational metrics, including model parameters, multiply-accumulate operations, inference throughput, training memory footprint, and template size, facilitates direct system comparison.
    
    \item Cross-spectral VIS–NIR verification protocols are incorporated as a robustness stress test, while the primary focus remains on efficient, reduced-input periocular recognition under real-world constraints.

    \item Systematic ablation studies evaluate the impact of model scaling across key architectural dimensions: backbone depth, channel width, embedding dimensionality, and spatial bottleneck design, revealing critical trade-offs between efficiency and performance.
\end{itemize}

The rest of the article is organized as follows: Section~\ref{sec:related_work} summarizes the state of the art. Section~\ref{sec:datasets} describes the set of images used, input conditions, and the metrics.  Section~\ref {sec:method} describes our proposal. The experimental framework of this work is then presented in Section~\ref{sec:expandresults}. The discussion is presented in Section~\ref {sec:discussion}, and the conclusion of the article in Section~\ref{sec:conclusion}.

\section{Related Work}
\label{sec:related_work}

Recent border-management technology roadmaps, such as those from European Border and Coast Guard Agency (Frontex), highlight "biometrics on the move" for future travel as an important operational topic \cite{frontex2023technology}. This initiative motivates the study of robust and efficient recognition methods for semi-control environments and partially visible biometric inputs.

Modern face recognition is dominated by deep convolutional embeddings trained on large amounts of data using angular-margin losses~\cite{kim2022adaface, meng2021magface}. One of the most widely used implementations is AdaFace~\cite{kim2022adaface}, which adapts the classification margin based on estimated image quality and provides strong baselines for face recognition. 

Cross-spectral face recognition is important for low-light and biometric-on-the-move applications. Recent lightweight approaches also study how to reduce model complexity under VIS--NIR domain shift~\cite{HFB,Anjith2026}.

Periocular recognition has been studied as a biometric alternative when the full face is partially unavailable, masked, or difficult to acquire, for example in semi-controlled, mobile, or biometric-on-the-move scenarios. 

Classical iris recognition relies on high-quality iris texture, segmentation, normalization, as well as specialized comparison algorithms~\cite{Daugman}. This is a different acquisition regime from the aligned face and periocular crops considered here. 

The periocular region is not defined by a single universally adopted standard and has no standardized extent. Prior work has noted that the term generally refers to the area around the eye, while exact boundaries vary across studies and datasets~\cite{Talreja2021ADPR}. We therefore explicitly define the direct periocular crop used in this paper instead of treating it as a standard crop (Sub-section:\ref{subsec:input}). 

Recent surveys frame the periocular region~\cite{AlonsoFernandez2024PeriocularUnconstrained, AlonsoFernandez2025OcularCropping, ALONSOFERNANDEZ2022110} as a useful cue under partial occlusion, mobile acquisition, unconstrained imaging as well as heterogeneous spectra~\cite{SharmaRoss2023PeriocularSurvey, AlonsoFernandez2024PeriocularUnconstrained}. Prior work on periocular verification and periocular datasets also shows that the eye region contains useful identity information under mobile and unconstrained conditions~\cite{Selfie-Tapia,zanlorensi2022new, Peri-PAD}. In contrast to work that primarily evaluates existing feature extractors or image enhancement methods, we focus on how much neural network capacity is needed once the input region itself is reduced.

Regarding the data, few high-resolution datasets are available, and some were captured several years ago~\cite{Ubiris, Visob}; as a result, available training data does not represent the real-world scenario in this domain. While being forced to rely on older training data, we can test on current datasets, like \textit{UFPR-periocular}~\cite{zanlorensi2022new}. It is a native periocular dataset and one of the latest collected in the state of the art.



\begin{table*}[t]
\centering
\scriptsize
\caption{Datasets and roles. Counts refer to the data version used in this work when available from the prepared manifests.}
\label{tab:datasets}
\begin{tabularx}{\textwidth}{lllr rY}
\toprule
Dataset & Role & Spectrum & IDs & Images & Notes \\
\midrule
MS1MV2 & training & VIS face & 85,164 & 3,804,846 & MTCNN aligned $112\times112$ used for all training runs \\
LFW, CFP-FP, AgeDB-30, CALFW, CPLFW & validation & VIS face & -- & 31,000 pairs & AdaFace model-selection validation set.\cite{kim2022adaface} \\
CASIA NIR-VIS 2.0 & test & VIS/NIR face & 725 & 17,345 & VIS--NIR final protocol \\
ND-NIVL & test & VIS/NIR face & 584 & 22,228 & VIS--NIR final protocol \\
FERET & test & VIS face & 994 & 8,564 & VIS--VIS final protocol \\
UFPR Periocular & test & VIS periocular & 1,122 & 16,789 & retained after preprocessing from $16,830$ native periocular images \\
\bottomrule
\end{tabularx}
\end{table*}

Regarding the model's efficiency, based on our review, there is no periocular recognition work that also focuses on the computational complexity of the models. Efficient model \cite{pmlr-v97-tan19a} design is commonly analyzed in terms of input resolution, network depth, channel width, and computational cost. 

The \textit{EfficientNet} \cite{pmlr-v97-tan19a} formalized the idea that these dimensions should be scaled jointly rather than independently. Recent ocular-cropping work has also trained CNNs on large-scale face-derived ocular crops and compared architectures such as \textit{SqueezeNet}, \textit{MobileNetV2} as well as \textit{ResNet50} \cite{AlonsoFernandez2025OcularCropping}. That line of work is close to our motivation, but it does not isolate full-face, masked-periocular as well as direct low-resolution periocular recognition within the same AdaFace-style scaling framework.

Efficiency-oriented periocular work also includes lightweight backbones, quantization, compression as well as mobile periocular authentication. We treat these directions as complementary. Our focus is not post-hoc compression or a new mobile architecture; instead, we ask what recognition-stage compute is saved when the input itself is reduced and the IR-style backbone, embedding size as well as final spatial representation are scaled under a common training protocol.

\section{Datasets}
\label{sec:datasets}

Several datasets were used in this work; the~\textit{MS1MV2} \cite{MS1MV2} dataset is used to train all models. The standard~\textit{AdaFace} validation sets are used for model selection~\cite{kim2022adaface}. The \textit{CASIA NIR-VIS} 2.0~\cite{li2013casia}, \textit{ND-NIVL}~\cite{ND-NIVL}, \textit{FERET}~\cite{PhillipsFRGC} as well as \textit{UFPR-Periocular}~\cite{zanlorensi2022new} are used for final verification benchmarks. 

The \textit{CASIA} and \textit{ND-NIVL} datasets are evaluated using VIS--NIR protocols, \textit{FERET} provides a VIS--VIS \textit{face protocol}, while \textit{UFPR} provides a \textit{periocular VIS protocol}.

The \textit{CASIA NIR-VIS 2.0} database offers pairs of mugshot images and their corresponding Near InfraRed (NIR) and Visible-Spectrum (VIS) images. The images in this database were collected in four recording sessions: 2007 spring, 2009 summer, 2009 fall, and 2010 summer, with the first session identical to the \textit{CASIA-HFB} database~\cite{HFB}. 

The \textit{ND-NIVL} dataset was collected by the University of Notre Dame, USA. The \textit{ND-NIVL} contains VIS and NIR face images from the same subjects. The capture process was carried out over two semesters (fall 2011 and spring 2012). The VIS images were collected using a \textit{Nikon-D90} camera. The \textit{Nikon-D90} uses a $23.6\times15.8$ mm CMOS sensor, and the resulting images have a resolution of $4288\times2848$ pixels. The images were acquired using automatic focus settings. All images were acquired under normal indoor lighting at a \textit{5-foot} stand-off, with a frontal pose and a neutral expression.

The NIR images were acquired using a \textit{Honeywell CFAIRS} system. The CFAIRS uses a modified \textit{Canon-EOS-50D} camera with a $22.3\times14.9$ CMOS sensor. The resulting images have a resolution of $4770\times3177$. All images were acquired under normal indoor lighting with frontal pose and neutral facial expression. NIR images were acquired at both a \textit{5-foot} and \textit{7-foot} stand-off. 

The \textit{UFPR-Periocular} dataset was created to capture images in uncontrolled environments, incorporating realistic noise from factors such as occlusion, blur, and variations in lighting, distance, and angles. To facilitate the capture process, was developed a mobile application that allows participants to collect their own pictures using their smartphones. The dataset contains $16,830$ images of both eyes from $1,122$ subjects. The resolution of the images varies, depending on the mobile devices used to capture the images.

\begin{figure*}[]
\centering
\includegraphics[width=2.8cm, height=3.4cm]{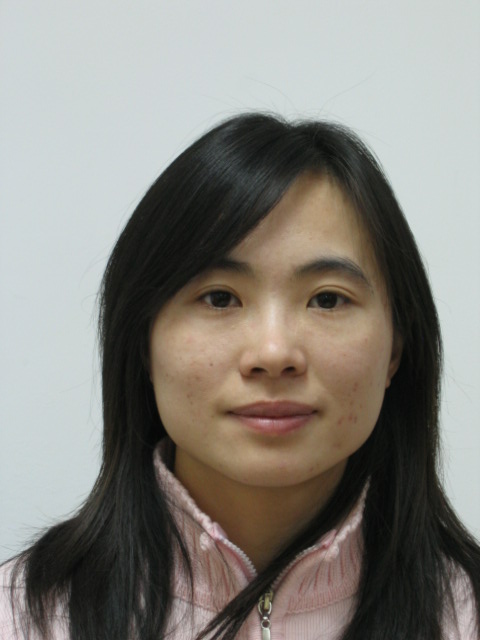}
\includegraphics[width=2.8cm, height=3.4cm]{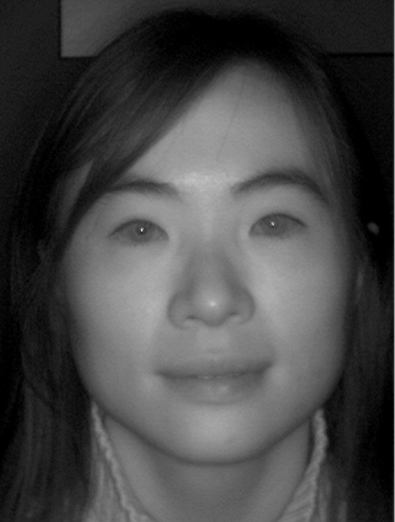}
\includegraphics[width=2.8cm, height=3.4cm]{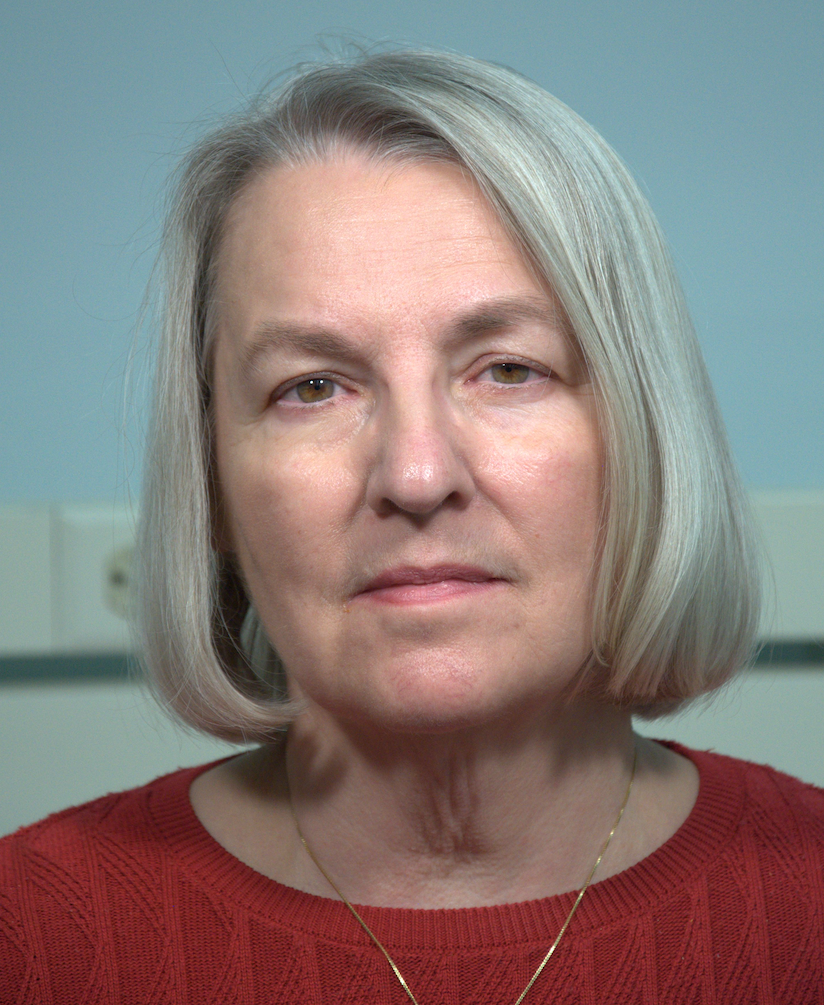}
\includegraphics[scale=0.078]{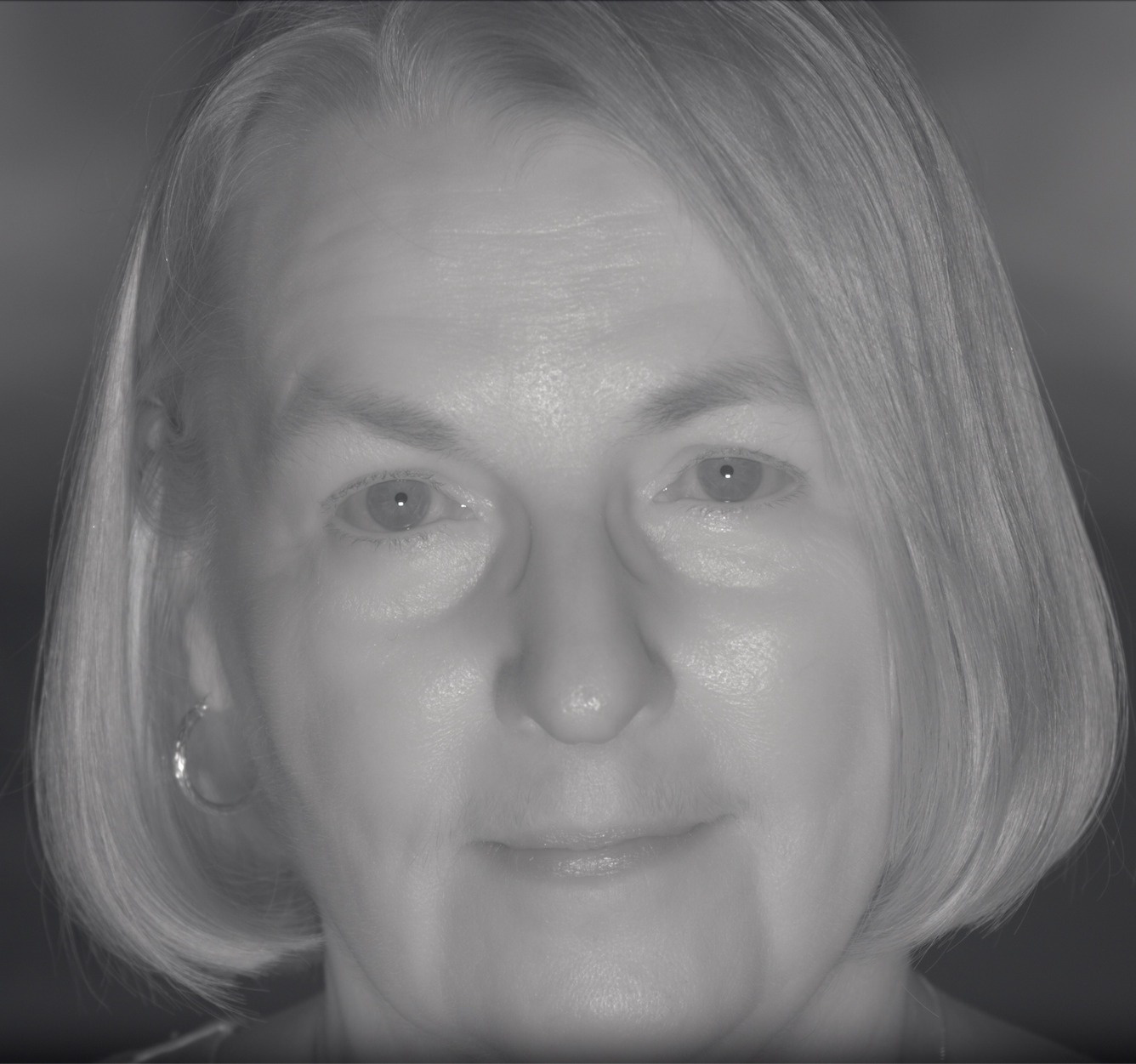}
\caption{\label{images-example} Left: Examples of the CASIA VIS-NIR dataset and Right: Examples of the ND-NIVL VIS-NIR dataset.}
\label{fig:vis_v1}
\end{figure*}

Table~\ref{tab:datasets} summarizes the datasets used in this work. Figure~\ref {images-example} illustrates an example of the datasets.

\subsection{Input Definitions}
\label{subsec:input}
All inputs for~\textit{AdaEye} follow the~\textit{AdaFace} preprocessing convention. \textit{MTCNN}~\cite{MTCNN} provides face detections and five landmark locations for alignment and cropping to $112\times112$. Pixel-value normalization is applied separately before network inference. No universally adopted periocular ROI definition fixes the crop location or aspect ratio. The crop was selected as a compact rectangular eye-region input that is compatible with the spatial dimensions of the IR-style backbone and avoids irregular final-layer dimensions. 

We explicitly define the direct crop as $y=27:67$, $x=16:96$ in the aligned $112\times112$ image, resulting in an $80\times40$ input. This compact rectangular crop preserves the eye-region context in the aligned coordinate system and yields regular feature-map dimensions for the IR-style backbone. 

For masked-periocular evaluation, the network input remains~$112\times112$, but all pixels outside the same periocular region are set to zero.

\noindent\textit{UFPR preprocessing.}~\textit{UFPR} contains native, tightly cropped periocular images rather than complete faces. Before detection, each image is centered on a gray canvas $(RGB 128, 128, 128)$, adding a border of width $p=\min(W,H)$ on all four sides, where $W$ and $H$ are the original image dimensions. Padding preserves the original pixels without resizing; the padded image is then processed by the detection and alignment pipeline. When the~\textit{MTCNN-based} path fails,~\textit{MediaPipe-FaceMesh}\cite{Lugaresi2019} provides a fallback set of five landmarks. These correspond to the two eyes, nose and two mouth corners; the five~\textit{MediaPipe} landmark indices were selected once by manual comparison with the~\textit{MTCNN} reference locations and remain fixed for all images. The fallback replaces landmark estimation while retaining the same five-point alignment procedure, reference template and $112\times112$ output size.

Historical processing logs show that MTCNN successfully processed 8,397 of 16,830 images (49.893\%). MediaPipe recovered a further 8,392 (49.863\%), yielding 16,789 usable outputs (99.756\%). The remaining 41 images (0.244\%) were excluded after both paths failed.

\begin{figure*}[t]
\centering
\begin{subfigure}{0.245\textwidth}
\centering
\parbox[c][0.43\textwidth][c]{\linewidth}{\centering\includegraphics[height=0.40\textwidth,keepaspectratio]{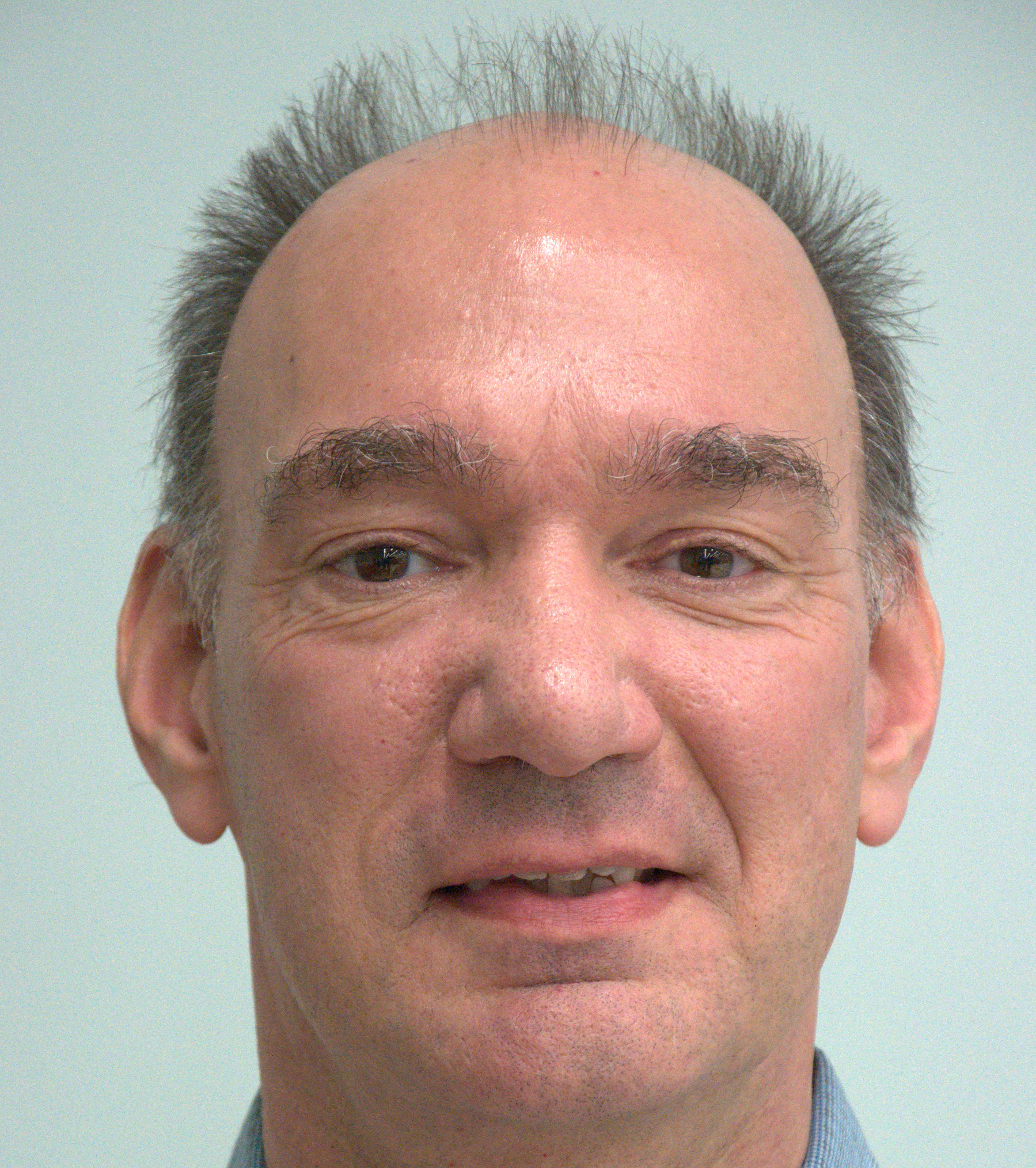}}
\caption{VIS original}
\end{subfigure}
\hfill
\begin{subfigure}{0.245\textwidth}
\centering
\parbox[c][0.43\textwidth][c]{\linewidth}{\centering\includegraphics[height=0.40\textwidth,keepaspectratio]{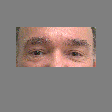}}
\caption{VIS periocular masked}
\end{subfigure}
\hfill
\begin{subfigure}{0.245\textwidth}
\centering
\parbox[c][0.43\textwidth][c]{\linewidth}{\centering\includegraphics[height=0.40\textwidth,keepaspectratio]{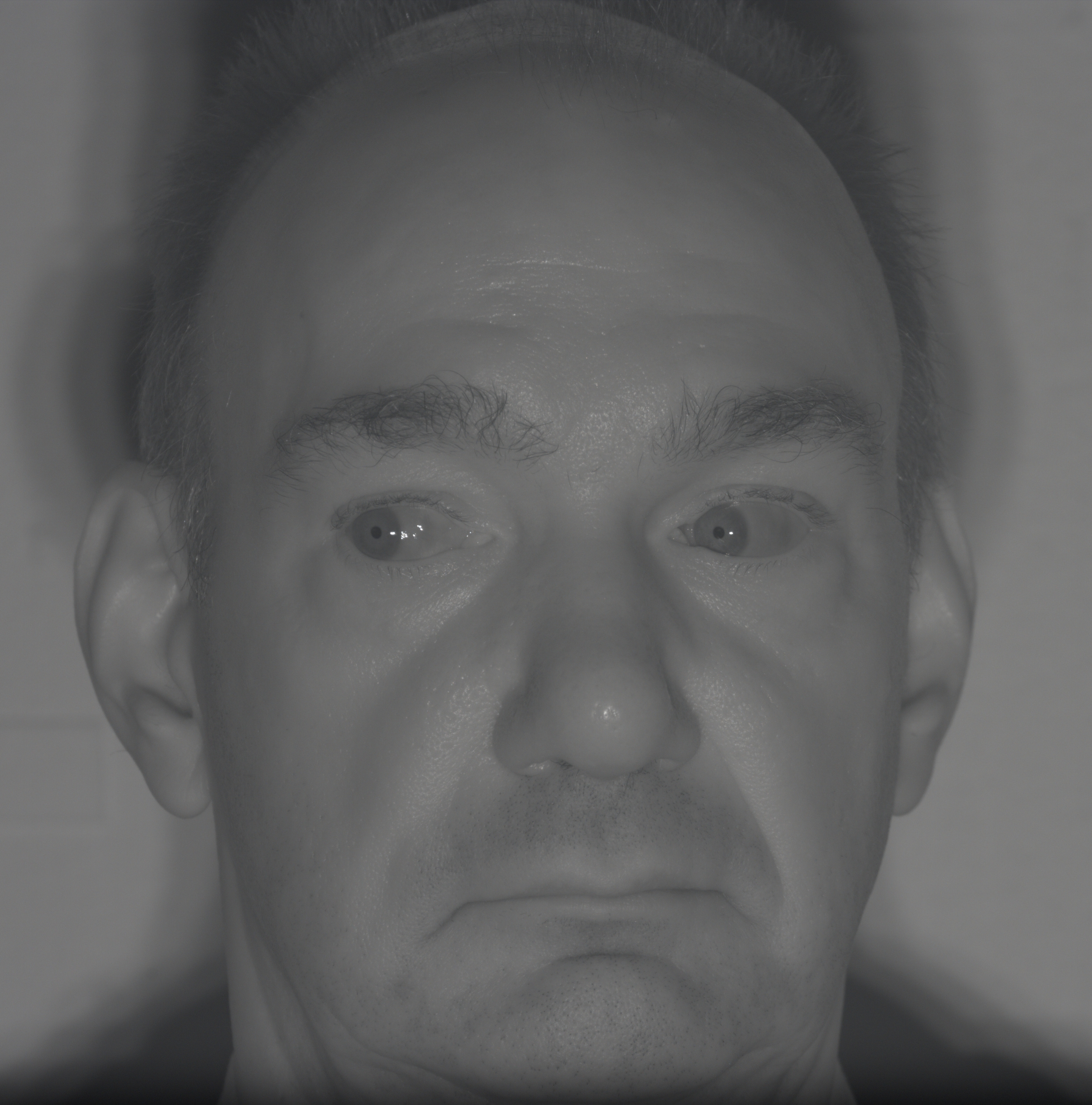}}
\caption{NIR original}
\end{subfigure}
\hfill
\begin{subfigure}{0.245\textwidth}
\centering
\parbox[c][0.43\textwidth][c]{\linewidth}{\centering\includegraphics[height=0.40\textwidth,keepaspectratio]{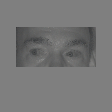}}
\caption{NIR periocular masked}
\end{subfigure}

\vspace{0.5em}

\begin{subfigure}{0.245\textwidth}
\centering
\parbox[c][0.30\textwidth][c]{\linewidth}{\centering\includegraphics[height=0.3\textwidth,keepaspectratio]{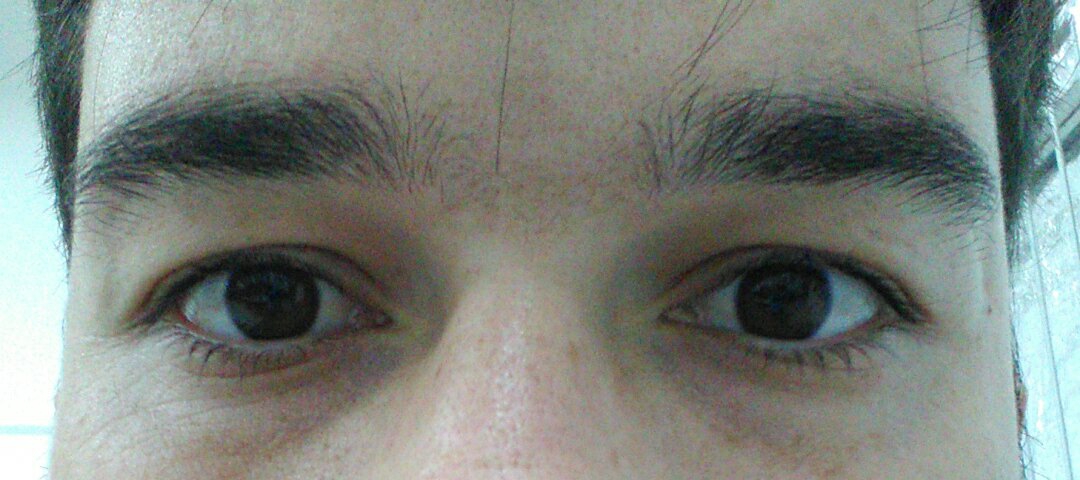}}
\caption{original}
\end{subfigure}
\hfill
\begin{subfigure}{0.245\textwidth}
\centering
\parbox[c][0.30\textwidth][c]{\linewidth}{\centering\includegraphics[height=0.3\textwidth,keepaspectratio]{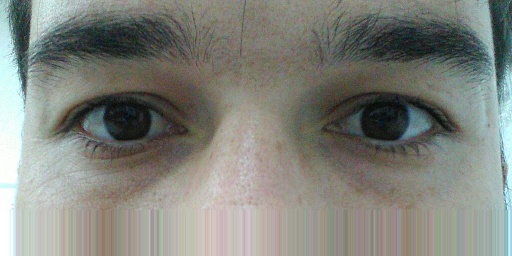}}
\caption{original aligned}
\end{subfigure}
\hfill
\begin{subfigure}{0.245\textwidth}
\centering
\parbox[c][0.30\textwidth][c]{\linewidth}{\centering\includegraphics[height=0.3\textwidth,keepaspectratio]{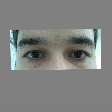}}
\caption{AdaFace-PP $112\times112$}
\end{subfigure}
\hfill
\begin{subfigure}{0.245\textwidth}
\centering
\parbox[c][0.30\textwidth][c]{\linewidth}{\centering\includegraphics[height=0.3\textwidth,keepaspectratio]{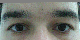}}
\caption{AdaFace-PP $80\times40$}
\end{subfigure}
\caption{Input construction examples. Top row: shows \textit{ND-NIVL} VIS and NIR aligned face images with their corresponding periocular crops. Bottom row: shows a native \textit{UFPR} periocular image, the \textit{UFPR} provided aligned version, the \textit{AdaFace} compatible $112\times112$ image obtained through padded detection and alignment, and the corresponding $80\times40$ periocular input.}
\label{fig:crop_examples}
\end{figure*}

All full-size, masked and direct-periocular evaluations use the same retained images and aligned $112\times112$ outputs; only the subsequent input definition differs. For \textit{UFPR}, ``full-face'' therefore denotes the full-size input condition, not the presence of a complete face. Figure~\ref{fig:crop_examples} illustrates the transformations for~\textit{UFPR} and the input definitions for~\textit{ND-NIVL}.

\subsection{Metrics}

Given a threshold $\tau$, we distinguish between mated and non-mated comparison pairs. Mated pairs correspond to comparisons between images of the same subject, while non-mated pairs correspond to comparisons between different subjects. For a distance-based score, a verification attempt is accepted if $s \leq \tau$ and rejected otherwise. The \emph{false non-match rate} (FNMR) at threshold $\tau$ is defined in Eq. \ref{eq:FNMR}.

\begin{equation}
\label{eq:FNMR}
    \mathrm{FNMR}(\tau)
    = \frac{\left|\{\text{mated comparisons with } s > \tau\}\right|}
           {\left|\{\text{all mated comparisons}\}\right|}
\end{equation}
that is, the proportion of mated comparisons that are incorrectly rejected.
Conversely, the \emph{false match rate} (FMR) at threshold $\tau$ is the proportion of non-mated comparisons that are incorrectly accepted, Eq. \ref{eq:FMR}. 

\begin{equation}
\label{eq:FMR}
    \mathrm{FMR}(\tau)
    = \frac{\left|\{\text{non-mated comparisons with } s \le \tau\}\right|}
           {\left|\{\text{all non-mated comparisons}\}\right|}
\end{equation}

Biometric performance is reported using Equal Error Rate (EER) and False Non-match Rate at fixed False Match Rate (\textit{FNMR@FMR}). The \textit{EER} summarizes the threshold-independent operating point at which the false match rate and the false non-match rate coincide. The \textit{FNMR@FMR} represents the biometric performance at an operationally relevant security level of maximum acceptance of non-mated comparison trials.

\section{Method}
\label{sec:method} 
We trained all versions from scratch and refer to those as \textit{\textbf{AdaEye}}. During training, each model state is saved and used for validation. For the main result tables, we select the model state that maximizes the unweighted mean accuracy over the standard AdaFace validation benchmarks \textit{LFW}~\cite{LFW}, \textit{CFP-FP}~\cite{CFPFP}, \textit{AgeDB-30}~\cite{AGEDB}, \textit{CALFW}~\cite{CALFW} as well as \textit{CPLFW }~\cite{CPLFW}.

Target evaluation protocols such as~\textit{CASIA},~\textit{ND-NIVL},~\textit{FERET} as well as~\textit{UFPR} are~\textit{\textbf{only used for testing}} and not take part of selection criterion. This avoids selecting a model specifically for a single final protocol or a single VIS--NIR dataset.

We use custom exhaustive verification protocols over the retained images, including all eligible mated and non-mated pairs without session-based restrictions or a separate gallery/probe partition; VIS–NIR protocols include only cross-spectral pairs. The same image inventories and pair lists are used for every model and training seed. Table~\ref{tab:protocols} summarizes the full protocol sizes. VIS--NIR are strictly cross-spectral and contain only VIS--NIR image pairs, not within-spectrum comparisons.

\begin{table}[H]
\centering
\scriptsize
\caption{Final verification protocols. All eligible mated and non-mated pairs are used.}
\label{tab:protocols}
\begin{tabular}{lrr}
\toprule
Protocol & Mated & Non-mated \\
\midrule
CASIA VIS--NIR & 87,478 & 62,297,636 \\
ND-NIVL VIS--NIR & 117,696 & 46,385,124 \\
FERET VIS & 57,189 & 36,609,577 \\
UFPR VIS & 117,260 & 140,809,606 \\
\bottomrule
\end{tabular}
\end{table}
\vspace{-0.2cm}

\subsection{Training protocol}
All variants and models are trained from scratch on \textit{MS1MV2 dataset} under the same optimization protocol: $60$ epochs, batch size $512$, \textit{SGD} with momentum $0.9$, initial learning rate $0.1$, \textit{MultiStepLR} decay by $0.1$ at epochs $18$, $30$ and $42$, mixed-precision training. The main biometric results aggregate successful runs from seeds $42$, $50$ and $51$. B1 and B2 reuse the corresponding A2 and A1 checkpoints for each seed.

Evaluation across seeds assesses both variation in verification performance and training reliability under the fixed training protocol. Biometric results are summarized as mean and sample standard deviation over successful runs among seeds 42, 50 and 51; the tables report the number of contributing runs, $n$. 

The~\textit{AdaFace} loss is used with the parameters $m=0.4$, $h=0.333$, $s=64$ and $t_\alpha=0.01$. The augmentation probabilities are $0.2$ for low-resolution cropping and photometric augmentation. For periocular variants, these augmentations are applied to the aligned $112\times112$ input before extracting the fixed periocular crop. Weight decay of $5\cdot10^{-4}$ is applied to convolutional, linear, and classification-head parameters, while batch-normalization parameters are excluded.

\subsection{Scaling Axes and Experiment Groups}
The experiments are organized using letters from~\textit{A} to \textit{K} to separate input-region effects from architectural capacity effects. \textit{C2} serves as the baseline for experiments \textit{E/F/G/H}. Table~\ref{tab:variants} lists all model identifiers, while this section defines the scaling axes used to construct them. This separation is important because the goal is not to propose a new backbone, but to quantify how much depth, channel width, embedding dimensionality, and late spatial capacity remain necessary once the input is restricted to the periocular region. Each number aside to the letter represents some change in the model compared with the previous one (i.e,~\textit{C1} and~\textit{C2}, the change is the backbone used).

\textit{\textbf{Input-region scaling}} compares three cases: 1) full-face $112\times112$ inputs, 2) masked-periocular $112\times112$ inputs and 3) direct $80\times40$ periocular inputs. 
The masked-periocular setting uses the same full-size network input as the full-face model but suppresses all pixels outside the fixed periocular ROI. It therefore measures whether a full-face model can exploit periocular information, but it does not reduce recognition-stage computation. The direct periocular setting crops the input area and reduces the input size.

\textit{\textbf{The channel-width}} scaling is implemented as a uniform backbone width multiplier. The multiplier is applied to the stem and all residual stages, followed by rounding to channel counts divisible by $8$. 

The residual block counts and spatial downsampling pattern remain unchanged, so width scaling affects representational channel capacity without changing depth or the final feature-map size. The resulting channel plans are given in Table~\ref{tab:width_scaling}.

\begin{table}[H]
\centering
\scriptsize
\caption{Backbone channel plans for width scaling. The multiplier is applied to the stem and all residual stages; channel counts are rounded to multiples of $8$.}
\label{tab:width_scaling}
\begin{tabular}{c|ccccc}
\toprule
Width & Stem & Stage 1 & Stage 2 & Stage 3 & Stage 4 \\
\midrule
1.00 & 64 & 64 & 128 & 256 & 512 \\
0.75 & 48 & 48 & 96 & 192 & 384 \\
0.50 & 32 & 32 & 64 & 128 & 256 \\
\bottomrule
\end{tabular}
\end{table}

\textit{\textbf{Depth scaling}} compares~\textit{IR-50} and~\textit{IR-18} backbones. The \textit{IR-50} variants use residual stage depths of $[3,4,14,3]$, while the~\textit{IR-18} variants use $[2,2,2,2]$. 
In both cases, the same stage layout is kept, so depth changes the number of residual units but not the input definition or recognition loss. 

\textit{\textbf{Embedding scaling}} changes the final embedding dimensionality from $512$ to $256$ or $128$. This directly reduces template size and the final projection layer, while leaving most convolutional computation unchanged. 

\textit{\textbf{Projections}}, for the default $80\times40$ architecture, the final feature map before the embedding projection is $3\times5$. This strongly reduces the final projection from~\textit{12.85M} parameters in the full-face~\textit{IR-50} reference to \textit{3.93M} parameters in the direct periocular~\textit{IR-50} baseline. 

\textit{\textbf{Spatial Representation}}: 
It disables the final downsampling step to retain a larger $5\times10$ map, increasing the final projection to \textit{13.11M} parameters and thereby testing whether the reduced, input-size-aligned periocular architecture creates an overly narrow final-projection bottleneck.

Moreover, it keeps this larger final map and uses dilation in the final stage to preserve a larger effective receptive field under the modified downsampling pattern. The~\textit{G} and~\textit{K} groups combine width and embedding reductions; \textit{K} is intentionally aggressive and is used to identify the convergence boundary of the fixed training protocol. Table \ref{tab:variants} summarizes all the experiment definitions ordered by identifier from~\textit{A} to~\textit{K}.

\begin{table*}[t]
\centering
\scriptsize
\setlength{\tabcolsep}{2pt}
\caption{Experiment definitions ordered by identifier. All models use the same \textit{AdaEye} training protocol unless indicated by the experiment identifier. Projection parameters refer to the final linear embedding projection.}
\label{tab:variants}
\begin{tabular}{lllcccllll}
\toprule
ID & Input & Resolution & Backbone & Width & Embedding & Final map & Stage4 & Proj. [M] & Notes \\
\midrule
A1 & full-face & $112\times112$ & IR-18 & 1.00 & 512 & $7\times7$ & default & 12.85 & \\
A2 & full-face & $112\times112$ & IR-50 & 1.00 & 512 & $7\times7$ & default & 12.85 & \\
\hline
B1 & masked periocular & $112\times112$ & IR-50 & 1.00 & 512 & $7\times7$ & default & 12.85 & A2 model weights, masked input \\
B2 & masked periocular & $112\times112$ & IR-18 & 1.00 & 512 & $7\times7$ & default & 12.85 & A1 model weights, masked input \\
\hline
C1 & periocular & $80\times40$ & IR-18 & 1.00 & 512 & $3\times5$ & default & 3.93 & \\
\mycc{C2} & \mycc{periocular} & \mycc{$80\times40$} & \mycc{IR-50} & \mycc{1.00} & \mycc{512} & \mycc{$3\times5$} & \mycc{default} & \mycc{3.93} & \mycc{baseline for E/F/G/H} \\
E2 & periocular & $80\times40$ & IR-50 & 0.75 & 512 & $3\times5$ & default & 2.95 & \\
E3 & periocular & $80\times40$ & IR-50 & 0.50 & 512 & $3\times5$ & default & 1.97 & \\
E4 & periocular & $80\times40$ & IR-18 & 0.75 & 512 & $3\times5$ & default & 2.95 & \\
E5 & periocular & $80\times40$ & IR-18 & 0.50 & 512 & $3\times5$ & default & 1.97 & \\
F2 & periocular & $80\times40$ & IR-50 & 1.00 & 256 & $3\times5$ & default & 1.97 & \\
F3 & periocular & $80\times40$ & IR-50 & 1.00 & 128 & $3\times5$ & default & 0.98 & \\
F4 & periocular & $80\times40$ & IR-18 & 1.00 & 256 & $3\times5$ & default & 1.97 & \\
F5 & periocular & $80\times40$ & IR-18 & 1.00 & 128 & $3\times5$ & default & 0.98 & \\
G1 & periocular & $80\times40$ & IR-50 & 0.75 & 256 & $3\times5$ & default & 1.47 & \\
G2 & periocular & $80\times40$ & IR-50 & 0.50 & 256 & $3\times5$ & default & 0.98 & \\
G4 & periocular & $80\times40$ & IR-18 & 0.75 & 256 & $3\times5$ & default & 1.47 & \\
G5 & periocular & $80\times40$ & IR-18 & 0.50 & 256 & $3\times5$ & default & 0.98 & \\
H2 & periocular & $80\times40$ & IR-50 & 1.00 & 512 & $5\times10$ & no downsample & 13.11 & \\
H3 & periocular & $80\times40$ & IR-50 & 1.00 & 512 & $5\times10$ & no downsample, dilated & 13.11 & \\
K2 & periocular & $80\times40$ & IR-50 & 0.75 & 128 & $3\times5$ & default & 0.74 & aggressive compression \\
K3 & periocular & $80\times40$ & IR-50 & 0.50 & 128 & $3\times5$ & default & 0.49 & aggressive compression \\
K4 & periocular & $80\times40$ & IR-18 & 0.75 & 128 & $3\times5$ & default & 0.74 & aggressive compression \\
K5 & periocular & $80\times40$ & IR-18 & 0.50 & 128 & $3\times5$ & default & 0.49 & aggressive compression \\
\bottomrule
\end{tabular}
\end{table*}

\begin{table*}[t]
\centering
\scriptsize
\setlength{\tabcolsep}{2pt}
\caption{Computational performance by experiment identifier. \textit{GMACs} and Images/s@512 refer to recognition-stage inference only. Inference parameters exclude the \textit{AdaEye} classification head, which is reported separately because it is used only during training. Peak train \textit{GB} is peak \textit{CUDA} allocated memory during synthetic batch-size-$512$ training steps. Template bytes assume uncompressed \textit{FP32} embedding size. \textit{C2} is the baseline.}
\label{tab:compute}
\begin{tabular}{lrrrrrr}
\toprule
ID & GMACs & Images/s@512 & Inf. params [M] & \textit{AdaEye} head [M] & Peak train [GB] & Template bytes \\
\midrule
A1 & 2.60 & 15,014 & 24.02 & 43.60 & 15.08 & 2,048 \\
A2 & 6.30 & 8,107 & 43.59 & 43.60 & 25.04 & 2,048 \\
\hline
B1 & 6.30 & 8,107 & 43.59 & 43.60 & 25.04 & 2,048 \\
B2 & 2.60 & 15,014 & 24.02 & 43.60 & 15.08 & 2,048 \\
\hline
C1 & 0.68 & 50,827 & 15.11 & 43.60 & 5.30 & 2,048 \\
\mycc{C2} & \mycc{1.64} & \mycc{25,295} & \mycc{34.67} & \mycc{43.60} & \mycc{7.94} & \mycc{2,048} \\
E2 & 0.92 & 32,905 & 20.25 & 43.60 & 6.40 & 2,048 \\
E3 & 0.41 & 50,062 & 9.66 & 43.60 & 4.88 & 2,048 \\
E4 & 0.38 & 65,463 & 9.24 & 43.60 & 4.43 & 2,048 \\
E5 & 0.17 & 101,485 & 4.76 & 43.60 & 3.58 & 2,048 \\
F2 & 1.63 & 25,266 & 32.71 & 21.80 & 7.81 & 1,024 \\
F3 & 1.63 & 25,259 & 31.72 & 10.90 & 7.73 & 512 \\
F4 & 0.68 & 51,206 & 13.14 & 21.80 & 5.17 & 1,024 \\
F5 & 0.68 & 51,202 & 12.16 & 10.90 & 5.10 & 512 \\
G1 & 0.92 & 33,129 & 18.77 & 21.80 & 6.27 & 1,024 \\
G2 & 0.41 & 44,010 & 8.68 & 21.80 & 7.00 & 1,024 \\
G4 & 0.38 & 65,313 & 7.76 & 21.80 & 4.31 & 1,024 \\
G5 & 0.17 & 101,561 & 3.78 & 21.80 & 3.45 & 1,024 \\
H2 & 2.06 & 23,497 & 43.85 & 43.60 & 8.41 & 2,048 \\
H3 & 2.06 & 23,843 & 43.85 & 43.60 & 8.41 & 2,048 \\
K2 & 0.92 & 33,163 & 18.04 & 10.90 & 6.20 & 512 \\
K3 & 0.41 & 50,213 & 8.19 & 10.90 & 4.68 & 512 \\
K4 & 0.38 & 65,293 & 7.03 & 10.90 & 4.24 & 512 \\
K5 & 0.17 & 101,629 & 3.29 & 10.90 & 3.39 & 512 \\
\bottomrule
\end{tabular}
\end{table*}

\section{Experiment and Results}
\label{sec:expandresults}

Inference throughput is measured on an \textit{NVIDIA H100 NVL} using~\textit{PyTorch 2.4.1},~\textit{CUDA 12.1} and automatic mixed precision where applicable. In Table \ref{tab:compute}, the column \textbf{\textit{Images/s@512}} denotes median inference throughput at batch size $512$ using synthetic tensors, model evaluation mode, no gradient tracking, and CUDA-event timing. Measurements use $100$ warm-up iterations followed by three measured repeats of $300$ iterations each. Dataloader, image decoding, detection, alignment and crop generation are excluded. 

\subsection{Computational performance}

Because computational performance is relevant to this study, we report Giga Multiply-Accumulate Operations (\textit{GMACs}) per image, inference throughput, inference parameter count,~\textit{AdaEye training-head size},~\textit{peak allocated training memory} and \textit{template size}. These values refer to the recognition network after successful detection, alignment, and~\textit{ROI} extraction. They do not include the upstream cost of face detection, landmark localization, padding-based fallback preprocessing, or crop generation. One~\textit{MAC} denotes one multiply-accumulate operation. 

The~\textit{\textbf{GMACs}} are counted for convolutional and linear layers of the inference embedding extractor and exclude the \textit{AdaEye} classification head used during training. The \textit{FLOP} counts are convention-dependent and would be twice the \textit{MAC} count under the common multiply-plus-add convention; we therefore report the \textit{GMACs} as the architecture-level compute measure. 

The \textit{\textbf{Parameter counts}} distinguish between the deployed recognition model and the training-only \textit{AdaEye} classification head. \textbf{\textit{Inference parameters}} include the backbone, optional post-body head and embedding projection. The \textbf{\textit{AdaEye head}} contains $d_{\mathrm{emb}}\times N_{\mathrm{cls}}$ parameters, where $d_{\mathrm{emb}}$ is the embedding dimensionality and $N_{\mathrm{cls}}=85{,}164$ for \textit{MS1MV2}. The total number of training parameters is therefore the sum of inference parameters and \textit{AdaEye head} parameters, but only the inference parameters remain in the deployed embedding extractor.

\textbf{\textit{Peak train GB}} reports peak CUDA allocated memory, not reserved caching-allocator memory, during synthetic batch-size-$512$ training steps with automatic mixed precision.  The training step includes the recognition model, \textit{AdaEye head}, loss, backward pass and optimizer update.

\textbf{\textit{Template bytes}} denote the raw uncompressed \textit{FP32} embedding payload, computed as $d_{\mathrm{emb}}\times4$ bytes, excluding identifiers, metadata, index structures, encryption, compression and file-system overhead.

Table~\ref{tab:compute} summarizes the computational profile of all model variants from \textit{A} to \textit{K}. The results show a trade-off among inference model size, GMACs, inference throughput, peak allocated training memory, and template size. Models with the highest throughput, such as \textit{E5} ($101{,}485$ images/s@512) and \textit{G5} ($101{,}561$ images/s@512), are also among the smallest in inference parameter count and computational demand. However, increasing capacity from \textit{E5} to \textit{A1} or \textit{H2} substantially reduces inference throughput, even when the template size remains unchanged.


\subsection{Accuracy-Efficiency Overview}
Tables~\ref{tab:casia_visnir}-\ref{tab:feret} report biometric performance based on \textit{EER} and~\textit{FNMR} at fixed \textit{FMR} operating points for each evaluation protocol. Full-face~\textit{IR50} remains the strongest VIS--NIR reference, but it is also the most computationally expensive model. 

Masked-periocular full-face baselines remain strong, yet they retain full $112\times112$ network cost. Direct $80\times40$ periocular models substantially reduce computation. The \textit{IR50} periocular baseline reduces the MAC count from $6.30$ \textit{GMACs} to $1.64$ \textit{GMACs} and increases inference throughput by approximately three times, while still retaining useful verification performance. The \textit{UFPR} protocol is particularly important because it is a native periocular dataset; here, the dedicated periocular baseline improves over the full-face reference. Checkpoints use the combined standard validation score.

\begin{table}[!t]
\centering
\scriptsize
\setlength{\tabcolsep}{2pt}
\renewcommand{\arraystretch}{1.05}
\caption{CASIA VIS--NIR biometric performance (\%). All eligible mated and non-mated pairs. Each cell shows the mean (top) and standard deviation (bottom) over $n$ successful runs; means unavailable for $n=1$. \textit{C2} is the baseline.}
\label{tab:casia_visnir}
\begin{tabular*}{\columnwidth}{@{\extracolsep{\fill}}lcrrrrr@{}}
\toprule
ID & $n$ & EER & \multicolumn{4}{c}{FNMR at FMR} \\
 & & & $10^{-1}$ & $10^{-2}$ & $10^{-3}$ & $10^{-4}$ \\
\midrule
A1 & 3 & \shortstack[r]{0.360\\{\tiny $\pm 0.018$}} & \shortstack[r]{0.052\\{\tiny $\pm 0.004$}} & \shortstack[r]{0.177\\{\tiny $\pm 0.015$}} & \shortstack[r]{0.799\\{\tiny $\pm 0.075$}} & \shortstack[r]{2.652\\{\tiny $\pm 0.173$}} \\
A2 & 3 & \shortstack[r]{0.117\\{\tiny $\pm 0.010$}} & \shortstack[r]{0.012\\{\tiny $\pm 0.002$}} & \shortstack[r]{0.026\\{\tiny $\pm 0.006$}} & \shortstack[r]{0.132\\{\tiny $\pm 0.021$}} & \shortstack[r]{0.697\\{\tiny $\pm 0.091$}} \\
\midrule
B1 & 3 & \shortstack[r]{1.576\\{\tiny $\pm 0.071$}} & \shortstack[r]{0.163\\{\tiny $\pm 0.026$}} & \shortstack[r]{2.442\\{\tiny $\pm 0.162$}} & \shortstack[r]{13.638\\{\tiny $\pm 0.764$}} & \shortstack[r]{37.656\\{\tiny $\pm 2.208$}} \\
B2 & 3 & \shortstack[r]{3.218\\{\tiny $\pm 0.160$}} & \shortstack[r]{0.814\\{\tiny $\pm 0.126$}} & \shortstack[r]{8.701\\{\tiny $\pm 0.334$}} & \shortstack[r]{29.771\\{\tiny $\pm 1.227$}} & \shortstack[r]{57.985\\{\tiny $\pm 1.646$}} \\
\midrule
C1 & 3 & \shortstack[r]{1.844\\{\tiny $\pm 0.061$}} & \shortstack[r]{0.362\\{\tiny $\pm 0.061$}} & \shortstack[r]{3.204\\{\tiny $\pm 0.121$}} & \shortstack[r]{14.108\\{\tiny $\pm 0.993$}} & \shortstack[r]{35.781\\{\tiny $\pm 0.822$}} \\
\mycc{C2} & 3 & \mycc{\shortstack[r]{1.358\\{\tiny $\pm 0.152$}}} & \mycc{\shortstack[r]{0.120\\{\tiny $\pm 0.027$}}} & \mycc{\shortstack[r]{1.872\\{\tiny $\pm 0.411$}}} & \mycc{\shortstack[r]{10.603\\{\tiny $\pm 1.453$}}} & \mycc{\shortstack[r]{28.651\\{\tiny $\pm 1.893$}}} \\
E2 & 3 & \shortstack[r]{1.283\\{\tiny $\pm 0.113$}} & \shortstack[r]{0.127\\{\tiny $\pm 0.015$}} & \shortstack[r]{1.658\\{\tiny $\pm 0.307$}} & \shortstack[r]{9.562\\{\tiny $\pm 1.151$}} & \shortstack[r]{28.690\\{\tiny $\pm 1.555$}} \\
E3 & 3 & \shortstack[r]{1.524\\{\tiny $\pm 0.113$}} & \shortstack[r]{0.194\\{\tiny $\pm 0.022$}} & \shortstack[r]{2.272\\{\tiny $\pm 0.243$}} & \shortstack[r]{11.101\\{\tiny $\pm 0.585$}} & \shortstack[r]{30.734\\{\tiny $\pm 1.424$}} \\
E4 & 3 & \shortstack[r]{2.332\\{\tiny $\pm 0.059$}} & \shortstack[r]{0.492\\{\tiny $\pm 0.023$}} & \shortstack[r]{4.657\\{\tiny $\pm 0.201$}} & \shortstack[r]{18.157\\{\tiny $\pm 2.067$}} & \shortstack[r]{43.215\\{\tiny $\pm 3.826$}} \\
E5 & 3 & \shortstack[r]{3.102\\{\tiny $\pm 0.200$}} & \shortstack[r]{0.934\\{\tiny $\pm 0.160$}} & \shortstack[r]{7.233\\{\tiny $\pm 0.966$}} & \shortstack[r]{25.109\\{\tiny $\pm 2.695$}} & \shortstack[r]{50.002\\{\tiny $\pm 4.071$}} \\
F2 & 1 & \shortstack[r]{1.424\\{\tiny --}} & \shortstack[r]{0.176\\{\tiny --}} & \shortstack[r]{1.884\\{\tiny --}} & \shortstack[r]{10.115\\{\tiny --}} & \shortstack[r]{28.787\\{\tiny --}} \\
F3 & 2 & \shortstack[r]{1.787\\{\tiny $\pm 0.090$}} & \shortstack[r]{0.251\\{\tiny $\pm 0.008$}} & \shortstack[r]{2.939\\{\tiny $\pm 0.339$}} & \shortstack[r]{13.717\\{\tiny $\pm 0.514$}} & \shortstack[r]{35.630\\{\tiny $\pm 1.591$}} \\
F4 & 3 & \shortstack[r]{1.813\\{\tiny $\pm 0.076$}} & \shortstack[r]{0.310\\{\tiny $\pm 0.023$}} & \shortstack[r]{3.118\\{\tiny $\pm 0.371$}} & \shortstack[r]{14.223\\{\tiny $\pm 0.691$}} & \shortstack[r]{36.026\\{\tiny $\pm 1.203$}} \\
F5 & 1 & \shortstack[r]{2.196\\{\tiny --}} & \shortstack[r]{0.351\\{\tiny --}} & \shortstack[r]{4.302\\{\tiny --}} & \shortstack[r]{18.365\\{\tiny --}} & \shortstack[r]{43.611\\{\tiny --}} \\
G1 & 1 & \shortstack[r]{1.205\\{\tiny --}} & \shortstack[r]{0.134\\{\tiny --}} & \shortstack[r]{1.515\\{\tiny --}} & \shortstack[r]{9.384\\{\tiny --}} & \shortstack[r]{27.794\\{\tiny --}} \\
G2 & 2 & \shortstack[r]{1.558\\{\tiny $\pm 0.008$}} & \shortstack[r]{0.198\\{\tiny $\pm 0.026$}} & \shortstack[r]{2.289\\{\tiny $\pm 0.027$}} & \shortstack[r]{11.587\\{\tiny $\pm 0.444$}} & \shortstack[r]{31.899\\{\tiny $\pm 0.992$}} \\
G4 & 2 & \shortstack[r]{2.230\\{\tiny $\pm 0.049$}} & \shortstack[r]{0.458\\{\tiny $\pm 0.012$}} & \shortstack[r]{4.151\\{\tiny $\pm 0.061$}} & \shortstack[r]{15.778\\{\tiny $\pm 1.179$}} & \shortstack[r]{38.186\\{\tiny $\pm 1.162$}} \\
G5 & 3 & \shortstack[r]{3.195\\{\tiny $\pm 0.101$}} & \shortstack[r]{1.033\\{\tiny $\pm 0.143$}} & \shortstack[r]{7.618\\{\tiny $\pm 0.584$}} & \shortstack[r]{25.674\\{\tiny $\pm 2.365$}} & \shortstack[r]{51.651\\{\tiny $\pm 3.965$}} \\
H2 & 3 & \shortstack[r]{1.201\\{\tiny $\pm 0.055$}} & \shortstack[r]{0.124\\{\tiny $\pm 0.025$}} & \shortstack[r]{1.438\\{\tiny $\pm 0.160$}} & \shortstack[r]{8.419\\{\tiny $\pm 0.865$}} & \shortstack[r]{25.206\\{\tiny $\pm 2.041$}} \\
H3 & 3 & \shortstack[r]{1.198\\{\tiny $\pm 0.139$}} & \shortstack[r]{0.138\\{\tiny $\pm 0.047$}} & \shortstack[r]{1.396\\{\tiny $\pm 0.286$}} & \shortstack[r]{8.017\\{\tiny $\pm 1.366$}} & \shortstack[r]{24.146\\{\tiny $\pm 3.834$}} \\
\midrule
\multicolumn{2}{l}{K2-K5} & \multicolumn{5}{c}{DVG (historical training failures)} \\
\bottomrule
\end{tabular*}
\end{table}

\begin{table}[!t]
\centering
\scriptsize
\setlength{\tabcolsep}{2pt}
\renewcommand{\arraystretch}{1.05}
\caption{ND-NIVL VIS--NIR biometric performance (\%). All eligible mated and non-mated pairs. Each cell shows the mean (top) and standard deviation (bottom) over $n$ successful runs; means unavailable for $n=1$. \textit{C2} is the baseline.}
\label{tab:nd_visnir}
\begin{tabular*}{\columnwidth}{@{\extracolsep{\fill}}lcrrrrr@{}}
\toprule
ID & $n$ & EER & \multicolumn{4}{c}{FNMR at FMR} \\
 & & & $10^{-1}$ & $10^{-2}$ & $10^{-3}$ & $10^{-4}$ \\
\midrule
A1 & 3 & \shortstack[r]{0.436\\{\tiny $\pm 0.020$}} & \shortstack[r]{0.304\\{\tiny $\pm 0.017$}} & \shortstack[r]{0.395\\{\tiny $\pm 0.019$}} & \shortstack[r]{0.530\\{\tiny $\pm 0.014$}} & \shortstack[r]{0.740\\{\tiny $\pm 0.016$}} \\
A2 & 3 & \shortstack[r]{0.324\\{\tiny $\pm 0.027$}} & \shortstack[r]{0.268\\{\tiny $\pm 0.030$}} & \shortstack[r]{0.307\\{\tiny $\pm 0.025$}} & \shortstack[r]{0.346\\{\tiny $\pm 0.042$}} & \shortstack[r]{0.407\\{\tiny $\pm 0.050$}} \\
\midrule
B1 & 3 & \shortstack[r]{0.787\\{\tiny $\pm 0.021$}} & \shortstack[r]{0.303\\{\tiny $\pm 0.027$}} & \shortstack[r]{0.688\\{\tiny $\pm 0.019$}} & \shortstack[r]{3.549\\{\tiny $\pm 0.367$}} & \shortstack[r]{14.811\\{\tiny $\pm 1.340$}} \\
B2 & 3 & \shortstack[r]{1.296\\{\tiny $\pm 0.127$}} & \shortstack[r]{0.420\\{\tiny $\pm 0.027$}} & \shortstack[r]{1.561\\{\tiny $\pm 0.258$}} & \shortstack[r]{7.491\\{\tiny $\pm 0.838$}} & \shortstack[r]{23.307\\{\tiny $\pm 1.675$}} \\
\midrule
C1 & 3 & \shortstack[r]{0.817\\{\tiny $\pm 0.030$}} & \shortstack[r]{0.332\\{\tiny $\pm 0.010$}} & \shortstack[r]{0.740\\{\tiny $\pm 0.039$}} & \shortstack[r]{3.108\\{\tiny $\pm 0.351$}} & \shortstack[r]{13.011\\{\tiny $\pm 1.474$}} \\
\mycc{C2} & 3 & \mycc{\shortstack[r]{0.697\\{\tiny $\pm 0.019$}}} & \mycc{\shortstack[r]{0.295\\{\tiny $\pm 0.026$}}} & \mycc{\shortstack[r]{0.580\\{\tiny $\pm 0.038$}}} & \mycc{\shortstack[r]{2.402\\{\tiny $\pm 0.030$}}} & \mycc{\shortstack[r]{9.482\\{\tiny $\pm 0.610$}}} \\
E2 & 3 & \shortstack[r]{0.687\\{\tiny $\pm 0.079$}} & \shortstack[r]{0.309\\{\tiny $\pm 0.017$}} & \shortstack[r]{0.590\\{\tiny $\pm 0.088$}} & \shortstack[r]{1.901\\{\tiny $\pm 0.301$}} & \shortstack[r]{8.526\\{\tiny $\pm 0.927$}} \\
E3 & 3 & \shortstack[r]{0.711\\{\tiny $\pm 0.056$}} & \shortstack[r]{0.311\\{\tiny $\pm 0.015$}} & \shortstack[r]{0.615\\{\tiny $\pm 0.052$}} & \shortstack[r]{2.342\\{\tiny $\pm 0.377$}} & \shortstack[r]{11.361\\{\tiny $\pm 1.299$}} \\
E4 & 3 & \shortstack[r]{1.026\\{\tiny $\pm 0.045$}} & \shortstack[r]{0.380\\{\tiny $\pm 0.020$}} & \shortstack[r]{1.040\\{\tiny $\pm 0.067$}} & \shortstack[r]{4.809\\{\tiny $\pm 0.118$}} & \shortstack[r]{15.647\\{\tiny $\pm 0.905$}} \\
E5 & 3 & \shortstack[r]{1.353\\{\tiny $\pm 0.161$}} & \shortstack[r]{0.470\\{\tiny $\pm 0.026$}} & \shortstack[r]{1.684\\{\tiny $\pm 0.342$}} & \shortstack[r]{7.219\\{\tiny $\pm 0.408$}} & \shortstack[r]{20.987\\{\tiny $\pm 1.904$}} \\
F2 & 1 & \shortstack[r]{0.692\\{\tiny --}} & \shortstack[r]{0.333\\{\tiny --}} & \shortstack[r]{0.597\\{\tiny --}} & \shortstack[r]{1.876\\{\tiny --}} & \shortstack[r]{7.800\\{\tiny --}} \\
F3 & 2 & \shortstack[r]{0.835\\{\tiny $\pm 0.041$}} & \shortstack[r]{0.277\\{\tiny $\pm 0.014$}} & \shortstack[r]{0.731\\{\tiny $\pm 0.088$}} & \shortstack[r]{3.508\\{\tiny $\pm 0.491$}} & \shortstack[r]{13.779\\{\tiny $\pm 2.145$}} \\
F4 & 3 & \shortstack[r]{0.871\\{\tiny $\pm 0.032$}} & \shortstack[r]{0.363\\{\tiny $\pm 0.051$}} & \shortstack[r]{0.814\\{\tiny $\pm 0.051$}} & \shortstack[r]{3.401\\{\tiny $\pm 0.228$}} & \shortstack[r]{14.081\\{\tiny $\pm 0.693$}} \\
F5 & 1 & \shortstack[r]{1.103\\{\tiny --}} & \shortstack[r]{0.325\\{\tiny --}} & \shortstack[r]{1.185\\{\tiny --}} & \shortstack[r]{5.916\\{\tiny --}} & \shortstack[r]{18.349\\{\tiny --}} \\
G1 & 1 & \shortstack[r]{0.693\\{\tiny --}} & \shortstack[r]{0.273\\{\tiny --}} & \shortstack[r]{0.593\\{\tiny --}} & \shortstack[r]{2.241\\{\tiny --}} & \shortstack[r]{9.086\\{\tiny --}} \\
G2 & 2 & \shortstack[r]{0.811\\{\tiny $\pm 0.084$}} & \shortstack[r]{0.329\\{\tiny $\pm 0.033$}} & \shortstack[r]{0.721\\{\tiny $\pm 0.110$}} & \shortstack[r]{3.140\\{\tiny $\pm 0.658$}} & \shortstack[r]{12.720\\{\tiny $\pm 2.471$}} \\
G4 & 2 & \shortstack[r]{0.992\\{\tiny $\pm 0.004$}} & \shortstack[r]{0.359\\{\tiny $\pm 0.025$}} & \shortstack[r]{0.988\\{\tiny $\pm 0.009$}} & \shortstack[r]{4.223\\{\tiny $\pm 0.049$}} & \shortstack[r]{15.385\\{\tiny $\pm 0.683$}} \\
G5 & 3 & \shortstack[r]{1.291\\{\tiny $\pm 0.108$}} & \shortstack[r]{0.457\\{\tiny $\pm 0.032$}} & \shortstack[r]{1.534\\{\tiny $\pm 0.221$}} & \shortstack[r]{6.689\\{\tiny $\pm 0.845$}} & \shortstack[r]{20.255\\{\tiny $\pm 0.837$}} \\
H2 & 3 & \shortstack[r]{0.631\\{\tiny $\pm 0.077$}} & \shortstack[r]{0.292\\{\tiny $\pm 0.008$}} & \shortstack[r]{0.533\\{\tiny $\pm 0.083$}} & \shortstack[r]{1.843\\{\tiny $\pm 0.285$}} & \shortstack[r]{8.471\\{\tiny $\pm 0.721$}} \\
H3 & 3 & \shortstack[r]{0.609\\{\tiny $\pm 0.050$}} & \shortstack[r]{0.290\\{\tiny $\pm 0.009$}} & \shortstack[r]{0.493\\{\tiny $\pm 0.043$}} & \shortstack[r]{1.670\\{\tiny $\pm 0.019$}} & \shortstack[r]{7.023\\{\tiny $\pm 0.610$}} \\
\midrule
\multicolumn{2}{l}{K2-K5} & \multicolumn{5}{c}{DVG (historical training failures)} \\
\bottomrule
\end{tabular*}
\end{table}

\begin{table}[!t]
\centering
\scriptsize
\setlength{\tabcolsep}{2pt}
\renewcommand{\arraystretch}{1.05}
\caption{UFPR biometric performance (\%). All eligible mated and non-mated pairs. Each cell shows the mean (top) and standard deviation (bottom) over $n$ successful runs; means unavailable for $n=1$. \textit{C2} is the baseline.}
\label{tab:ufpr}
\begin{tabular*}{\columnwidth}{@{\extracolsep{\fill}}lcrrrrr@{}}
\toprule
ID & $n$ & EER & \multicolumn{4}{c}{FNMR at FMR} \\
 & & & $10^{-1}$ & $10^{-2}$ & $10^{-3}$ & $10^{-4}$ \\
\midrule
A1 & 3 & \shortstack[r]{5.376\\{\tiny $\pm 0.169$}} & \shortstack[r]{3.375\\{\tiny $\pm 0.201$}} & \shortstack[r]{12.523\\{\tiny $\pm 0.715$}} & \shortstack[r]{25.826\\{\tiny $\pm 1.212$}} & \shortstack[r]{40.635\\{\tiny $\pm 1.535$}} \\
A2 & 3 & \shortstack[r]{3.083\\{\tiny $\pm 0.214$}} & \shortstack[r]{1.219\\{\tiny $\pm 0.138$}} & \shortstack[r]{6.025\\{\tiny $\pm 0.686$}} & \shortstack[r]{15.502\\{\tiny $\pm 1.548$}} & \shortstack[r]{28.708\\{\tiny $\pm 2.432$}} \\
\midrule
B1 & 3 & \shortstack[r]{3.075\\{\tiny $\pm 0.094$}} & \shortstack[r]{1.245\\{\tiny $\pm 0.022$}} & \shortstack[r]{5.910\\{\tiny $\pm 0.315$}} & \shortstack[r]{15.082\\{\tiny $\pm 0.590$}} & \shortstack[r]{27.931\\{\tiny $\pm 1.086$}} \\
B2 & 3 & \shortstack[r]{5.252\\{\tiny $\pm 0.116$}} & \shortstack[r]{3.339\\{\tiny $\pm 0.158$}} & \shortstack[r]{12.035\\{\tiny $\pm 0.433$}} & \shortstack[r]{24.896\\{\tiny $\pm 0.794$}} & \shortstack[r]{39.731\\{\tiny $\pm 1.156$}} \\
\midrule
C1 & 3 & \shortstack[r]{3.956\\{\tiny $\pm 0.126$}} & \shortstack[r]{2.116\\{\tiny $\pm 0.083$}} & \shortstack[r]{7.943\\{\tiny $\pm 0.419$}} & \shortstack[r]{17.132\\{\tiny $\pm 0.862$}} & \shortstack[r]{28.954\\{\tiny $\pm 0.965$}} \\
\mycc{C2} & 3 & \mycc{\shortstack[r]{2.020\\{\tiny $\pm 0.083$}}} & \mycc{\shortstack[r]{0.761\\{\tiny $\pm 0.040$}}} & \mycc{\shortstack[r]{2.969\\{\tiny $\pm 0.183$}}} & \mycc{\shortstack[r]{7.920\\{\tiny $\pm 0.570$}}} & \mycc{\shortstack[r]{16.213\\{\tiny $\pm 1.282$}}} \\
E2 & 3 & \shortstack[r]{2.505\\{\tiny $\pm 0.051$}} & \shortstack[r]{1.016\\{\tiny $\pm 0.050$}} & \shortstack[r]{4.014\\{\tiny $\pm 0.156$}} & \shortstack[r]{9.974\\{\tiny $\pm 0.426$}} & \shortstack[r]{19.071\\{\tiny $\pm 0.613$}} \\
E3 & 3 & \shortstack[r]{3.114\\{\tiny $\pm 0.059$}} & \shortstack[r]{1.442\\{\tiny $\pm 0.038$}} & \shortstack[r]{5.565\\{\tiny $\pm 0.130$}} & \shortstack[r]{12.995\\{\tiny $\pm 0.395$}} & \shortstack[r]{23.481\\{\tiny $\pm 0.631$}} \\
E4 & 3 & \shortstack[r]{4.715\\{\tiny $\pm 0.114$}} & \shortstack[r]{2.813\\{\tiny $\pm 0.150$}} & \shortstack[r]{10.093\\{\tiny $\pm 0.235$}} & \shortstack[r]{21.123\\{\tiny $\pm 0.411$}} & \shortstack[r]{34.327\\{\tiny $\pm 0.619$}} \\
E5 & 3 & \shortstack[r]{5.800\\{\tiny $\pm 0.072$}} & \shortstack[r]{4.035\\{\tiny $\pm 0.157$}} & \shortstack[r]{13.332\\{\tiny $\pm 0.175$}} & \shortstack[r]{25.922\\{\tiny $\pm 0.240$}} & \shortstack[r]{39.616\\{\tiny $\pm 0.199$}} \\
F2 & 1 & \shortstack[r]{2.256\\{\tiny --}} & \shortstack[r]{0.861\\{\tiny --}} & \shortstack[r]{3.470\\{\tiny --}} & \shortstack[r]{8.805\\{\tiny --}} & \shortstack[r]{17.865\\{\tiny --}} \\
F3 & 2 & \shortstack[r]{2.688\\{\tiny $\pm 0.191$}} & \shortstack[r]{1.135\\{\tiny $\pm 0.165$}} & \shortstack[r]{4.490\\{\tiny $\pm 0.420$}} & \shortstack[r]{11.065\\{\tiny $\pm 0.595$}} & \shortstack[r]{21.184\\{\tiny $\pm 0.620$}} \\
F4 & 3 & \shortstack[r]{4.083\\{\tiny $\pm 0.163$}} & \shortstack[r]{2.214\\{\tiny $\pm 0.123$}} & \shortstack[r]{8.308\\{\tiny $\pm 0.573$}} & \shortstack[r]{18.182\\{\tiny $\pm 1.108$}} & \shortstack[r]{30.849\\{\tiny $\pm 1.341$}} \\
F5 & 1 & \shortstack[r]{4.923\\{\tiny --}} & \shortstack[r]{3.114\\{\tiny --}} & \shortstack[r]{10.372\\{\tiny --}} & \shortstack[r]{20.989\\{\tiny --}} & \shortstack[r]{33.411\\{\tiny --}} \\
G1 & 1 & \shortstack[r]{2.416\\{\tiny --}} & \shortstack[r]{0.959\\{\tiny --}} & \shortstack[r]{3.813\\{\tiny --}} & \shortstack[r]{9.693\\{\tiny --}} & \shortstack[r]{18.913\\{\tiny --}} \\
G2 & 2 & \shortstack[r]{3.156\\{\tiny $\pm 0.186$}} & \shortstack[r]{1.513\\{\tiny $\pm 0.112$}} & \shortstack[r]{5.603\\{\tiny $\pm 0.482$}} & \shortstack[r]{12.897\\{\tiny $\pm 1.106$}} & \shortstack[r]{22.885\\{\tiny $\pm 1.292$}} \\
G4 & 2 & \shortstack[r]{4.709\\{\tiny $\pm 0.159$}} & \shortstack[r]{2.798\\{\tiny $\pm 0.124$}} & \shortstack[r]{9.938\\{\tiny $\pm 0.523$}} & \shortstack[r]{20.568\\{\tiny $\pm 0.539$}} & \shortstack[r]{33.473\\{\tiny $\pm 0.392$}} \\
G5 & 3 & \shortstack[r]{5.979\\{\tiny $\pm 0.243$}} & \shortstack[r]{4.240\\{\tiny $\pm 0.308$}} & \shortstack[r]{13.864\\{\tiny $\pm 0.980$}} & \shortstack[r]{26.678\\{\tiny $\pm 1.487$}} & \shortstack[r]{40.400\\{\tiny $\pm 1.736$}} \\
H2 & 3 & \shortstack[r]{2.075\\{\tiny $\pm 0.057$}} & \shortstack[r]{0.729\\{\tiny $\pm 0.047$}} & \shortstack[r]{3.096\\{\tiny $\pm 0.126$}} & \shortstack[r]{8.272\\{\tiny $\pm 0.270$}} & \shortstack[r]{16.633\\{\tiny $\pm 0.626$}} \\
H3 & 3 & \shortstack[r]{2.044\\{\tiny $\pm 0.050$}} & \shortstack[r]{0.723\\{\tiny $\pm 0.045$}} & \shortstack[r]{3.015\\{\tiny $\pm 0.152$}} & \shortstack[r]{7.973\\{\tiny $\pm 0.369$}} & \shortstack[r]{16.058\\{\tiny $\pm 0.554$}} \\
\midrule
\multicolumn{2}{l}{K2--K5} & \multicolumn{5}{c}{DVG (historical training failures)} \\
\bottomrule
\end{tabular*}
\end{table}

\begin{table}[!t]
\centering
\scriptsize
\setlength{\tabcolsep}{2pt}
\renewcommand{\arraystretch}{1.05}
\caption{FERET VIS--VIS biometric performance (\%). All eligible mated and non-mated pairs. Each cell shows the mean (top) and standard deviation (bottom) over $n$ successful runs; means unavailable for $n=1$. \textit{C2} is the baseline.}
\label{tab:feret}
\begin{tabular*}{\columnwidth}{@{\extracolsep{\fill}}lcrrrrr@{}}
\toprule
ID & $n$ & EER & \multicolumn{4}{c}{FNMR at FMR} \\
 & & & $10^{-1}$ & $10^{-2}$ & $10^{-3}$ & $10^{-4}$ \\
\midrule
A1 & 3 & \shortstack[r]{1.708\\{\tiny $\pm 0.042$}} & \shortstack[r]{1.054\\{\tiny $\pm 0.050$}} & \shortstack[r]{1.980\\{\tiny $\pm 0.077$}} & \shortstack[r]{3.824\\{\tiny $\pm 0.255$}} & \shortstack[r]{7.117\\{\tiny $\pm 0.369$}} \\
A2 & 3 & \shortstack[r]{1.051\\{\tiny $\pm 0.014$}} & \shortstack[r]{0.881\\{\tiny $\pm 0.057$}} & \shortstack[r]{1.053\\{\tiny $\pm 0.015$}} & \shortstack[r]{1.246\\{\tiny $\pm 0.019$}} & \shortstack[r]{1.792\\{\tiny $\pm 0.126$}} \\
\midrule
B1 & 3 & \shortstack[r]{2.530\\{\tiny $\pm 0.055$}} & \shortstack[r]{1.627\\{\tiny $\pm 0.045$}} & \shortstack[r]{3.484\\{\tiny $\pm 0.084$}} & \shortstack[r]{8.120\\{\tiny $\pm 0.122$}} & \shortstack[r]{17.349\\{\tiny $\pm 0.802$}} \\
B2 & 3 & \shortstack[r]{5.062\\{\tiny $\pm 0.291$}} & \shortstack[r]{3.615\\{\tiny $\pm 0.353$}} & \shortstack[r]{9.497\\{\tiny $\pm 0.434$}} & \shortstack[r]{19.412\\{\tiny $\pm 0.479$}} & \shortstack[r]{32.113\\{\tiny $\pm 0.721$}} \\
\midrule
C1 & 3 & \shortstack[r]{4.920\\{\tiny $\pm 0.297$}} & \shortstack[r]{3.694\\{\tiny $\pm 0.275$}} & \shortstack[r]{8.598\\{\tiny $\pm 0.837$}} & \shortstack[r]{20.832\\{\tiny $\pm 3.784$}} & \shortstack[r]{74.410\\{\tiny $\pm 7.314$}} \\
\mycc{C2} & 3 & \mycc{\shortstack[r]{2.880\\{\tiny $\pm 0.019$}}} & \mycc{\shortstack[r]{1.946\\{\tiny $\pm 0.016$}}} & \mycc{\shortstack[r]{3.961\\{\tiny $\pm 0.132$}}} & \mycc{\shortstack[r]{7.530\\{\tiny $\pm 0.491$}}} & \mycc{\shortstack[r]{22.933\\{\tiny $\pm 4.216$}}} \\
E2 & 3 & \shortstack[r]{3.273\\{\tiny $\pm 0.122$}} & \shortstack[r]{2.294\\{\tiny $\pm 0.117$}} & \shortstack[r]{4.548\\{\tiny $\pm 0.233$}} & \shortstack[r]{8.604\\{\tiny $\pm 0.424$}} & \shortstack[r]{53.365\\{\tiny $\pm 7.397$}} \\
E3 & 3 & \shortstack[r]{3.827\\{\tiny $\pm 0.506$}} & \shortstack[r]{2.851\\{\tiny $\pm 0.587$}} & \shortstack[r]{5.563\\{\tiny $\pm 0.696$}} & \shortstack[r]{11.147\\{\tiny $\pm 1.551$}} & \shortstack[r]{55.664\\{\tiny $\pm 30.611$}} \\
E4 & 3 & \shortstack[r]{6.214\\{\tiny $\pm 0.344$}} & \shortstack[r]{5.037\\{\tiny $\pm 0.377$}} & \shortstack[r]{12.187\\{\tiny $\pm 0.649$}} & \shortstack[r]{38.824\\{\tiny $\pm 9.134$}} & \shortstack[r]{87.276\\{\tiny $\pm 6.536$}} \\
E5 & 3 & \shortstack[r]{8.645\\{\tiny $\pm 0.867$}} & \shortstack[r]{8.041\\{\tiny $\pm 1.222$}} & \shortstack[r]{24.221\\{\tiny $\pm 5.118$}} & \shortstack[r]{69.065\\{\tiny $\pm 8.307$}} & \shortstack[r]{94.386\\{\tiny $\pm 2.973$}} \\
F2 & 1 & \shortstack[r]{2.681\\{\tiny --}} & \shortstack[r]{1.944\\{\tiny --}} & \shortstack[r]{3.634\\{\tiny --}} & \shortstack[r]{6.828\\{\tiny --}} & \shortstack[r]{18.086\\{\tiny --}} \\
F3 & 2 & \shortstack[r]{3.203\\{\tiny $\pm 0.112$}} & \shortstack[r]{2.069\\{\tiny $\pm 0.176$}} & \shortstack[r]{4.634\\{\tiny $\pm 0.116$}} & \shortstack[r]{8.892\\{\tiny $\pm 0.135$}} & \shortstack[r]{17.182\\{\tiny $\pm 1.296$}} \\
F4 & 3 & \shortstack[r]{5.049\\{\tiny $\pm 0.290$}} & \shortstack[r]{3.863\\{\tiny $\pm 0.245$}} & \shortstack[r]{8.952\\{\tiny $\pm 0.858$}} & \shortstack[r]{23.620\\{\tiny $\pm 6.619$}} & \shortstack[r]{80.093\\{\tiny $\pm 8.652$}} \\
F5 & 1 & \shortstack[r]{5.123\\{\tiny --}} & \shortstack[r]{3.885\\{\tiny --}} & \shortstack[r]{9.329\\{\tiny --}} & \shortstack[r]{18.402\\{\tiny --}} & \shortstack[r]{41.386\\{\tiny --}} \\
G1 & 1 & \shortstack[r]{2.834\\{\tiny --}} & \shortstack[r]{1.897\\{\tiny --}} & \shortstack[r]{3.982\\{\tiny --}} & \shortstack[r]{8.145\\{\tiny --}} & \shortstack[r]{19.479\\{\tiny --}} \\
G2 & 2 & \shortstack[r]{3.786\\{\tiny $\pm 0.028$}} & \shortstack[r]{2.694\\{\tiny $\pm 0.070$}} & \shortstack[r]{5.665\\{\tiny $\pm 0.006$}} & \shortstack[r]{11.537\\{\tiny $\pm 0.094$}} & \shortstack[r]{58.464\\{\tiny $\pm 7.854$}} \\
G4 & 2 & \shortstack[r]{5.985\\{\tiny $\pm 0.211$}} & \shortstack[r]{4.829\\{\tiny $\pm 0.206$}} & \shortstack[r]{11.426\\{\tiny $\pm 0.842$}} & \shortstack[r]{37.528\\{\tiny $\pm 5.440$}} & \shortstack[r]{87.627\\{\tiny $\pm 0.905$}} \\
G5 & 3 & \shortstack[r]{8.371\\{\tiny $\pm 0.573$}} & \shortstack[r]{7.631\\{\tiny $\pm 0.824$}} & \shortstack[r]{20.967\\{\tiny $\pm 1.686$}} & \shortstack[r]{58.308\\{\tiny $\pm 14.088$}} & \shortstack[r]{86.269\\{\tiny $\pm 13.226$}} \\
H2 & 3 & \shortstack[r]{2.788\\{\tiny $\pm 0.081$}} & \shortstack[r]{1.930\\{\tiny $\pm 0.030$}} & \shortstack[r]{3.762\\{\tiny $\pm 0.143$}} & \shortstack[r]{7.538\\{\tiny $\pm 0.283$}} & \shortstack[r]{20.446\\{\tiny $\pm 3.689$}} \\
H3 & 3 & \shortstack[r]{2.866\\{\tiny $\pm 0.084$}} & \shortstack[r]{2.043\\{\tiny $\pm 0.112$}} & \shortstack[r]{3.790\\{\tiny $\pm 0.109$}} & \shortstack[r]{7.311\\{\tiny $\pm 0.257$}} & \shortstack[r]{21.547\\{\tiny $\pm 4.547$}} \\
\midrule
\multicolumn{2}{l}{K2--K5} & \multicolumn{5}{c}{DVG (historical training failures)} \\
\bottomrule
\end{tabular*}
\end{table}

\begin{figure*}[t]
\centering
\begin{subfigure}{0.48\textwidth}
\centering
\includegraphics[scale=0.35]{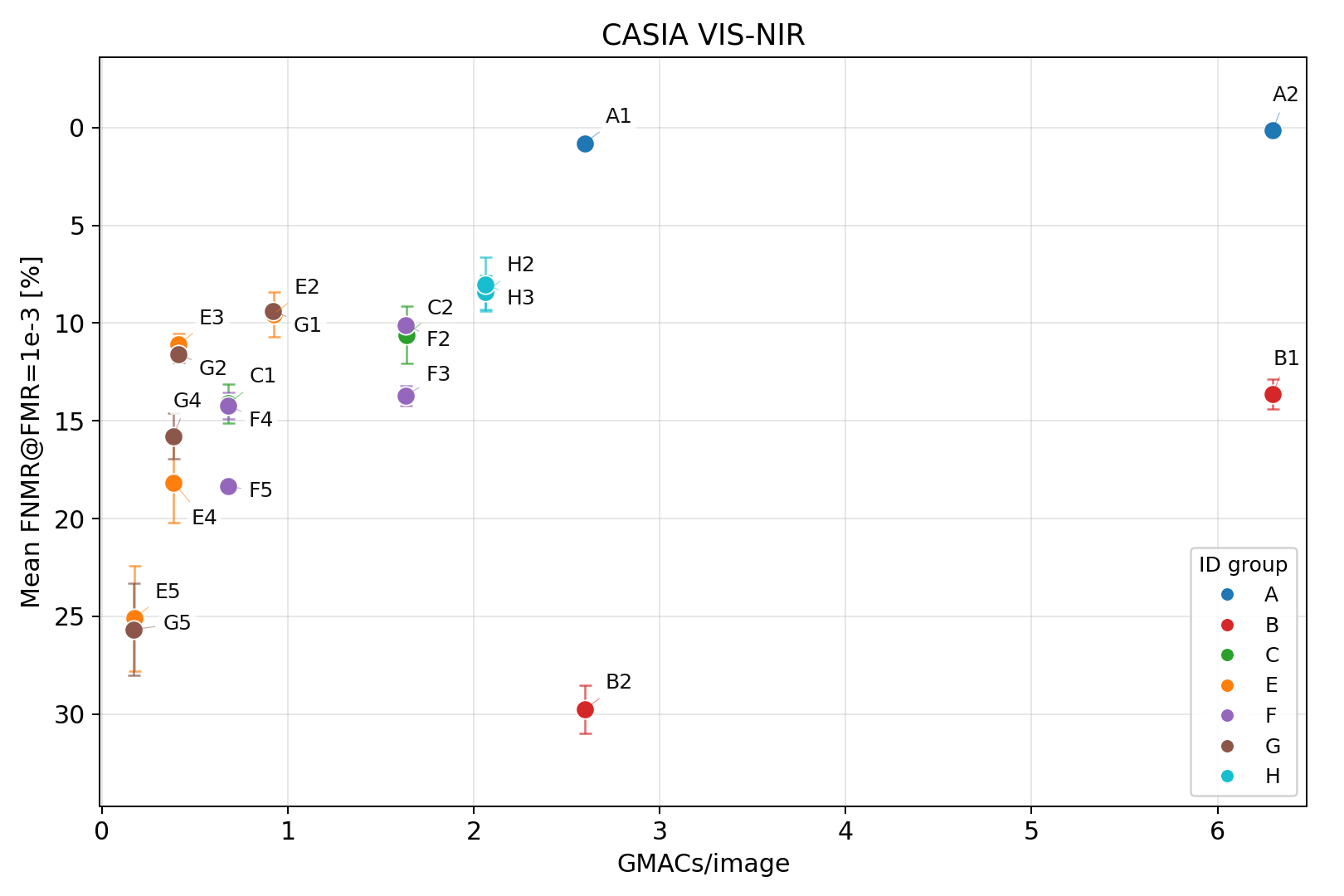}
\end{subfigure}
\hfill
\begin{subfigure}{0.48\textwidth}
\centering
\includegraphics[scale=0.35]{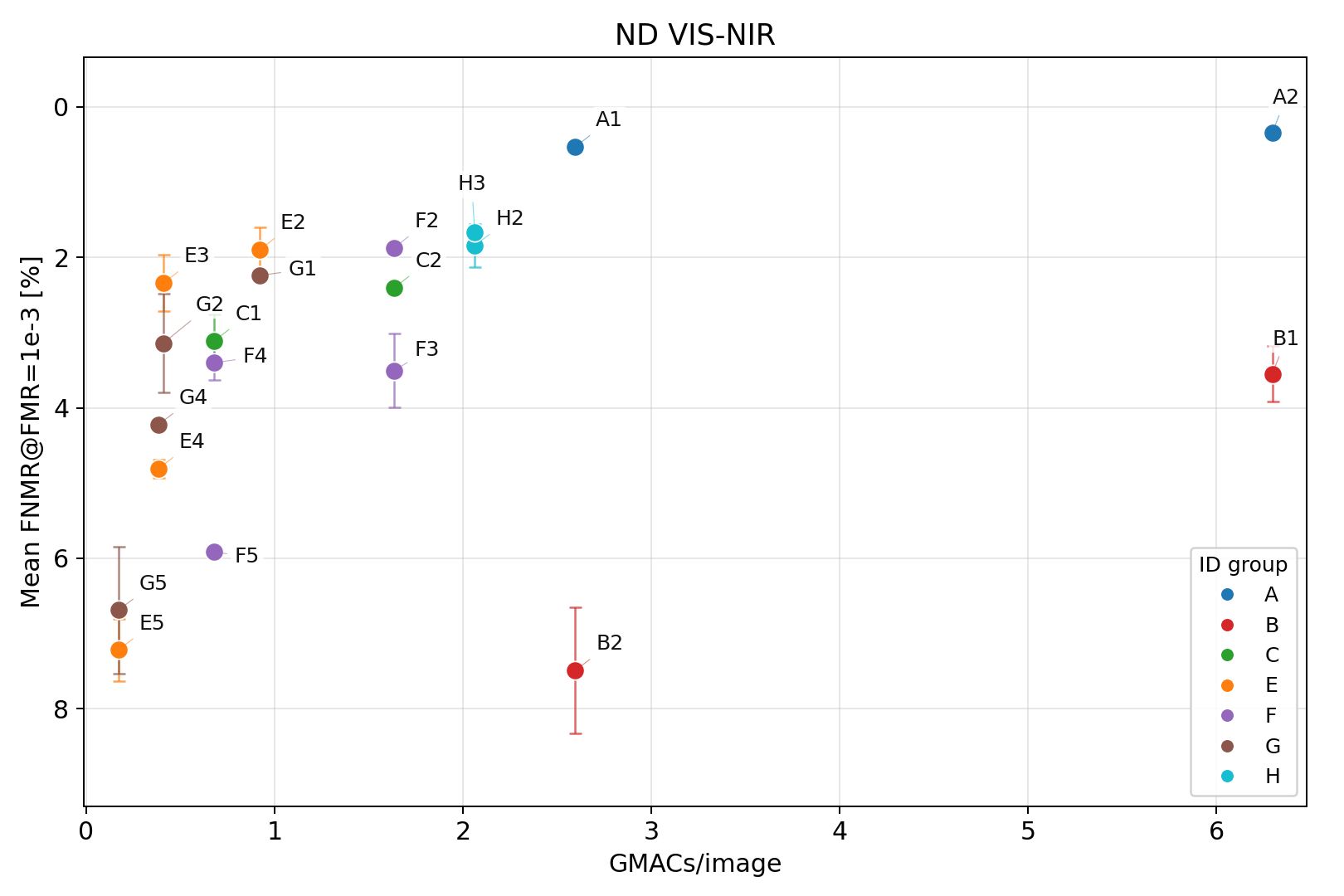}
\end{subfigure}

\vspace{0.6em}

\begin{subfigure}{0.48\textwidth}
\centering
\includegraphics[scale=0.35]{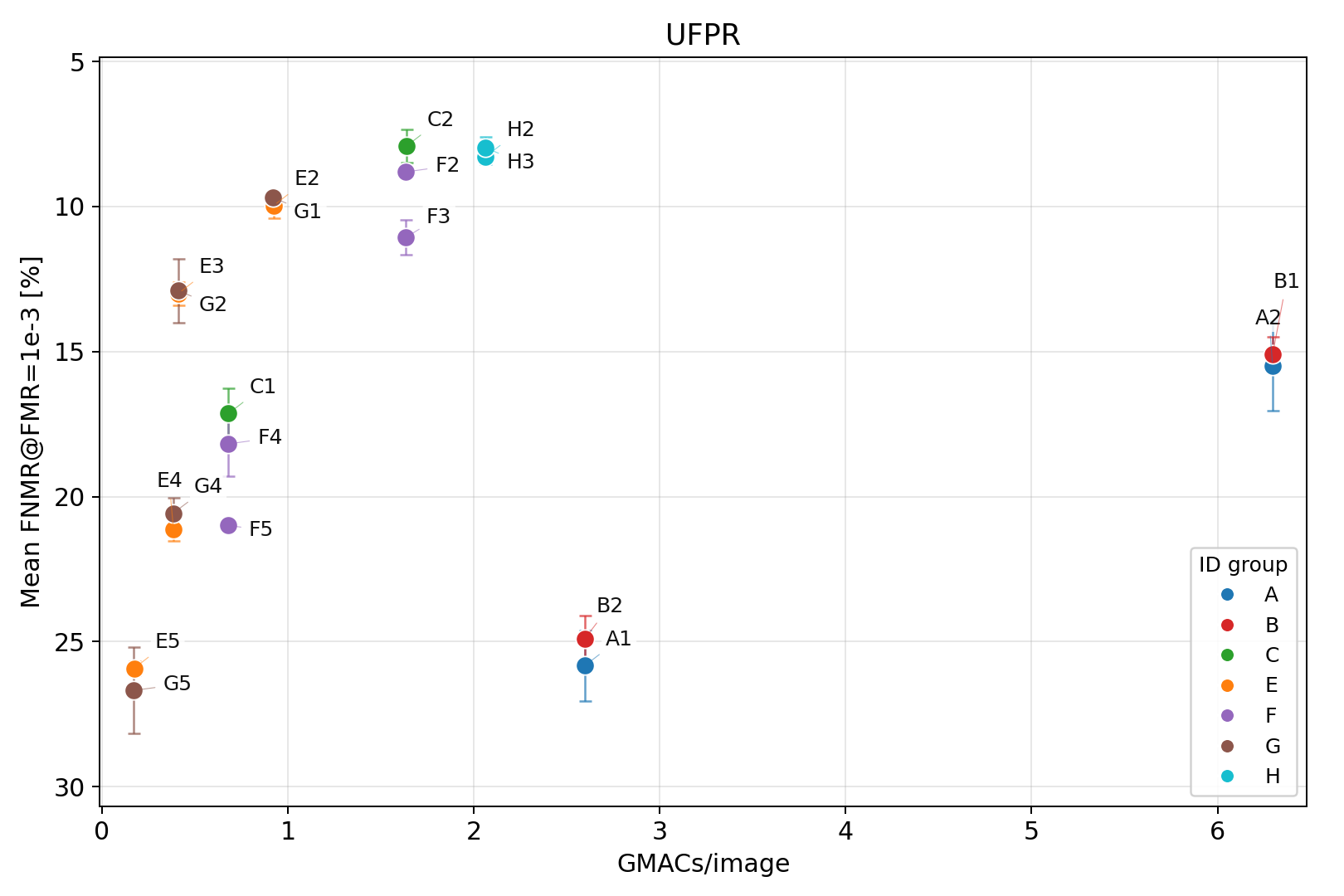}
\end{subfigure}
\hfill
\begin{subfigure}{0.48\textwidth}
\centering
\includegraphics[scale=0.35]{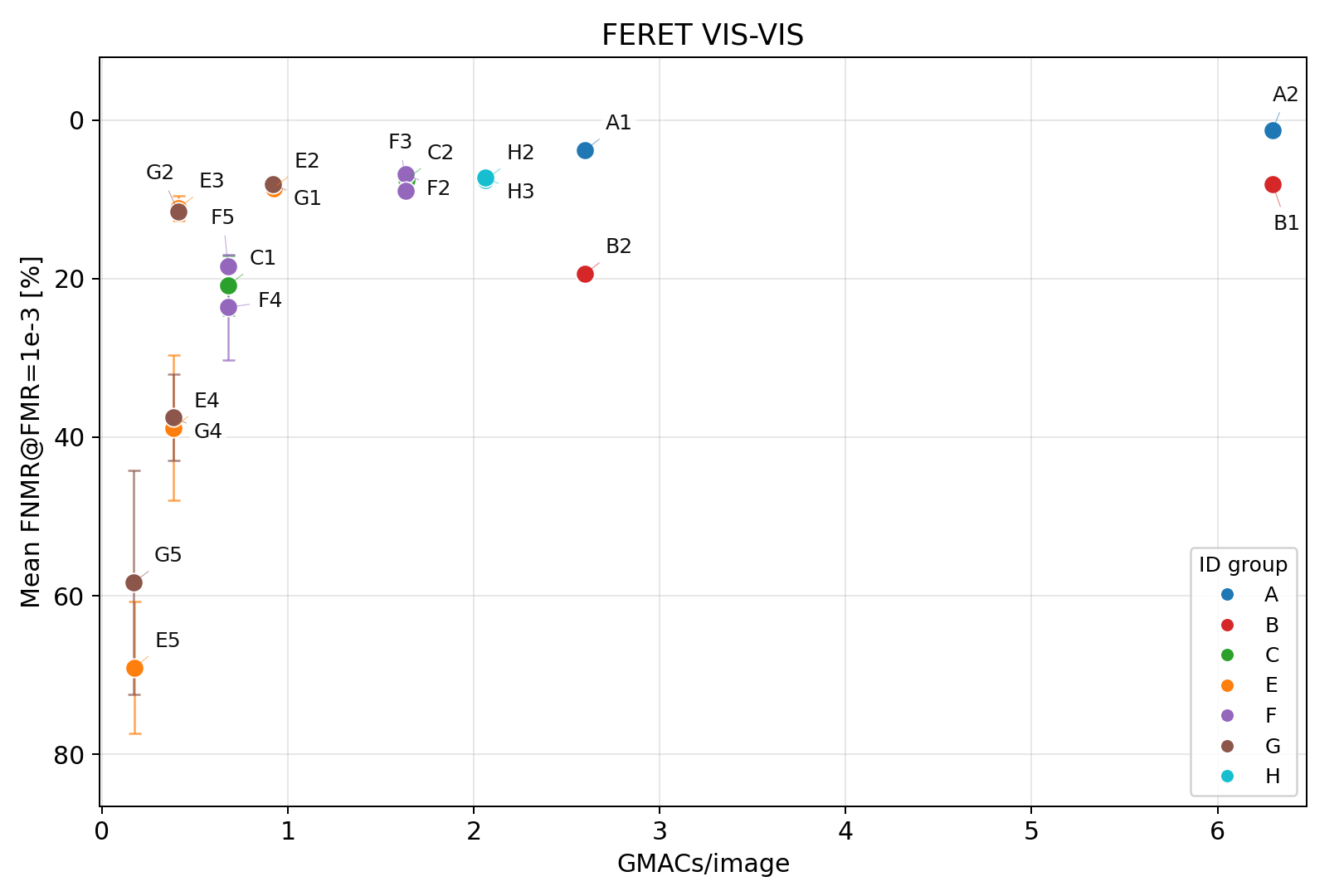}
\end{subfigure}
\caption{FNMR@FMR=$10^{-3}$ versus computational cost. Lower \textit{FNMR} and lower \textit{GMACs} are better. The plots emphasize operational behavior at a fixed low-FMR point and show how direct periocular models compare with masked full-face reuse.}
\label{fig:fnmr1e3_vs_gmacs}
\end{figure*}

\begin{figure*}[t]
\centering
\begin{subfigure}{0.48\textwidth}
\centering
\includegraphics[scale=0.35]{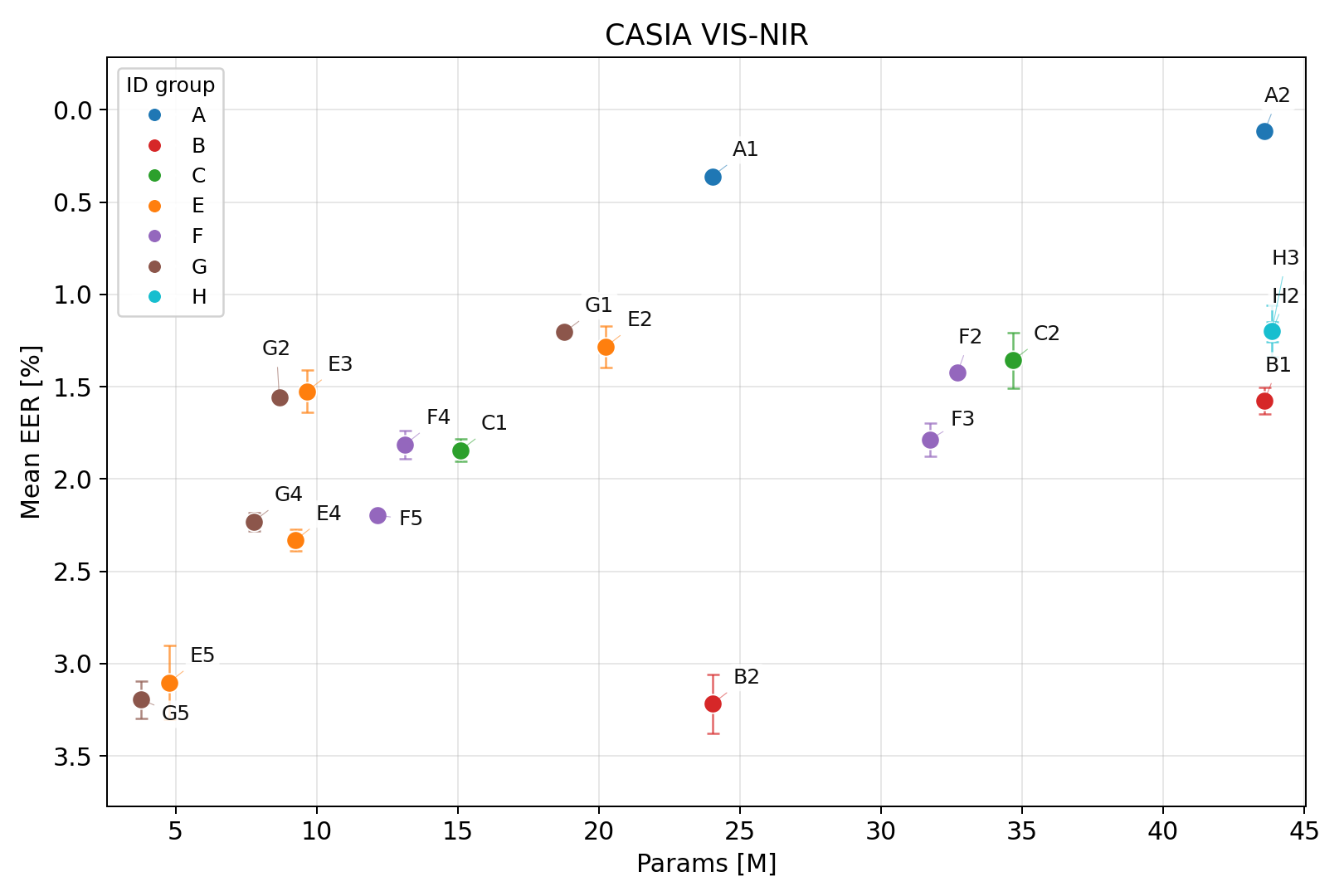}
\end{subfigure}
\hfill
\begin{subfigure}{0.48\textwidth}
\centering
\includegraphics[scale=0.35]{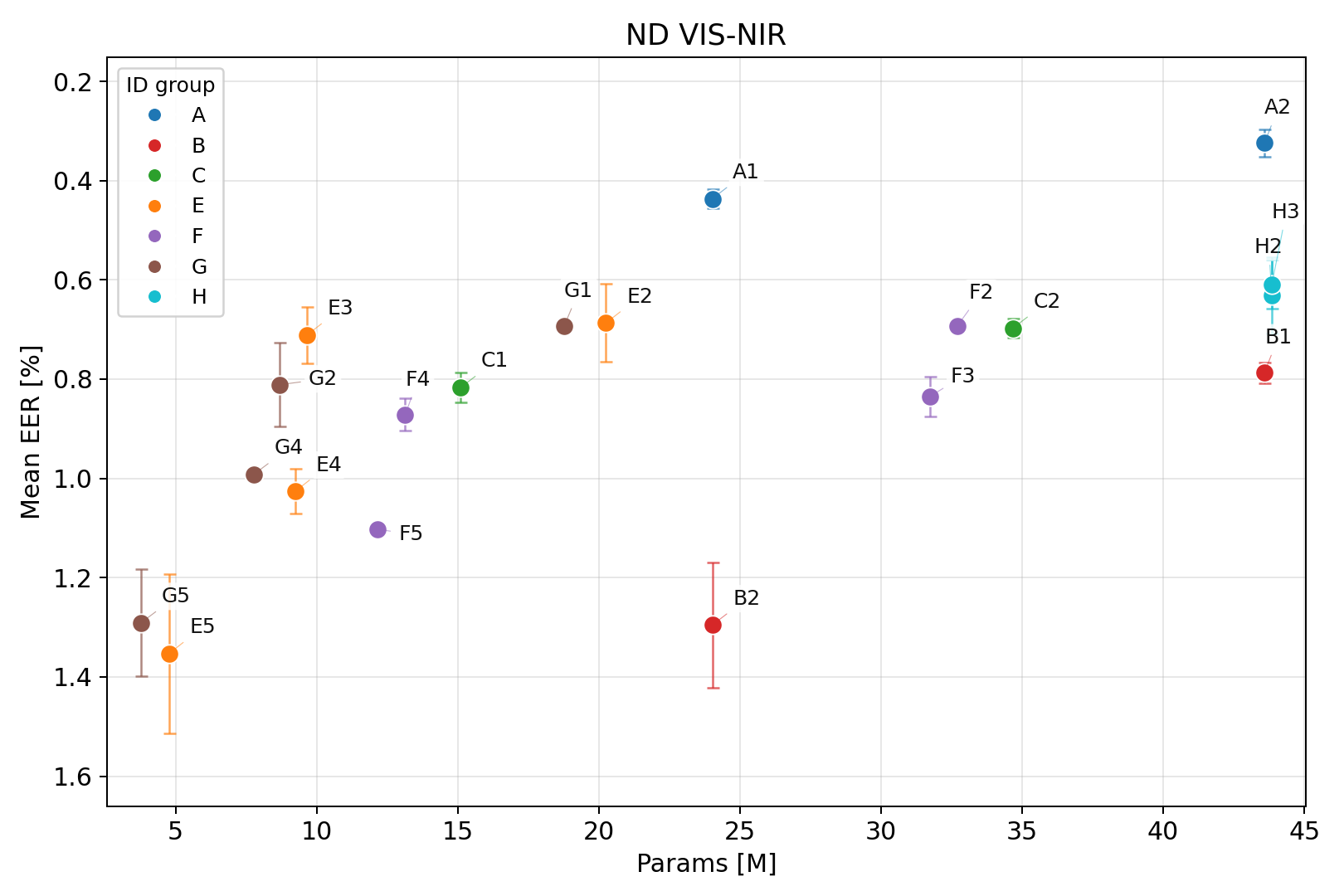}
\end{subfigure}

\vspace{0.6em}

\begin{subfigure}{0.48\textwidth}
\centering
\includegraphics[scale=0.35]{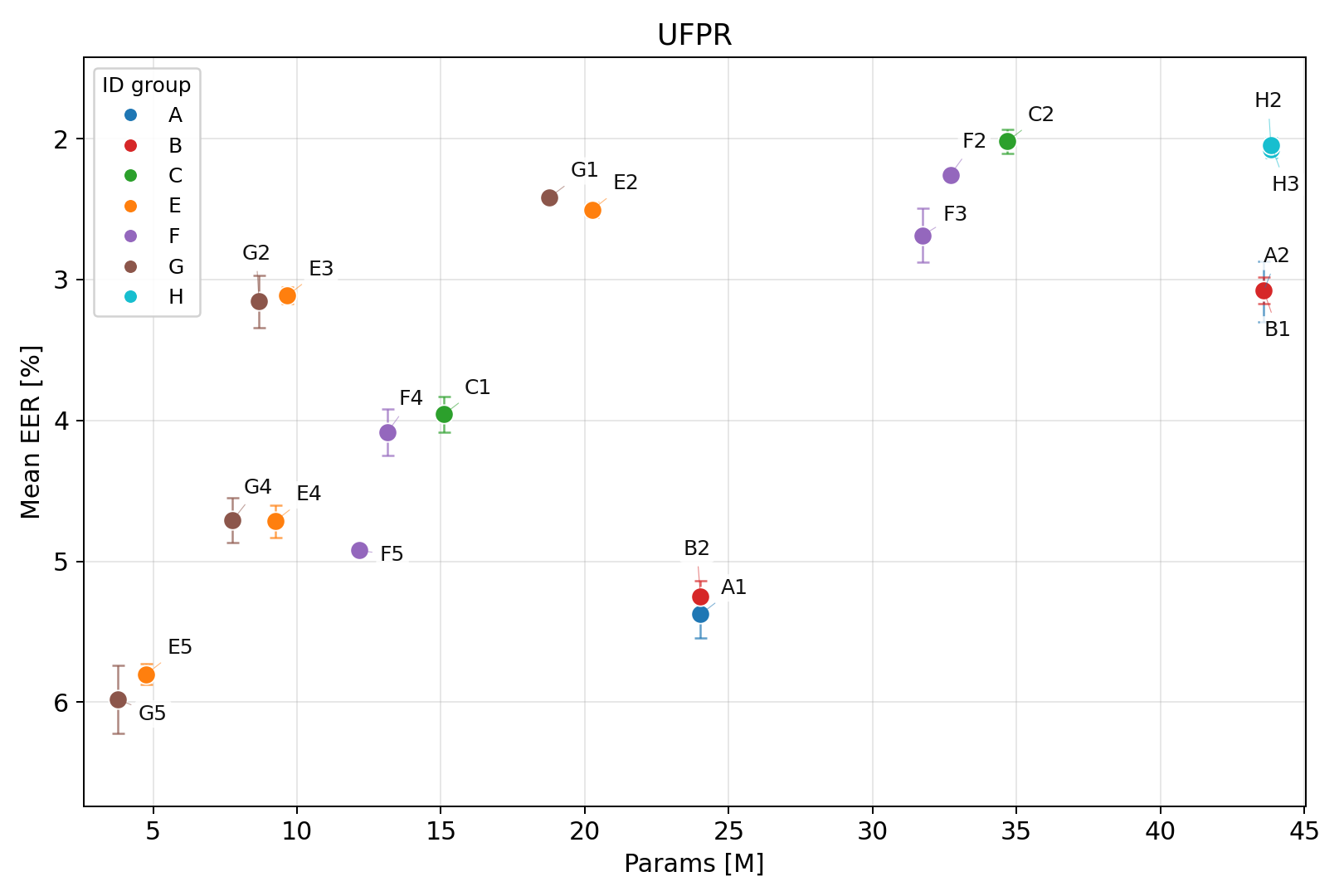}
\end{subfigure}
\hfill
\begin{subfigure}{0.48\textwidth}
\centering
\includegraphics[scale=0.35]{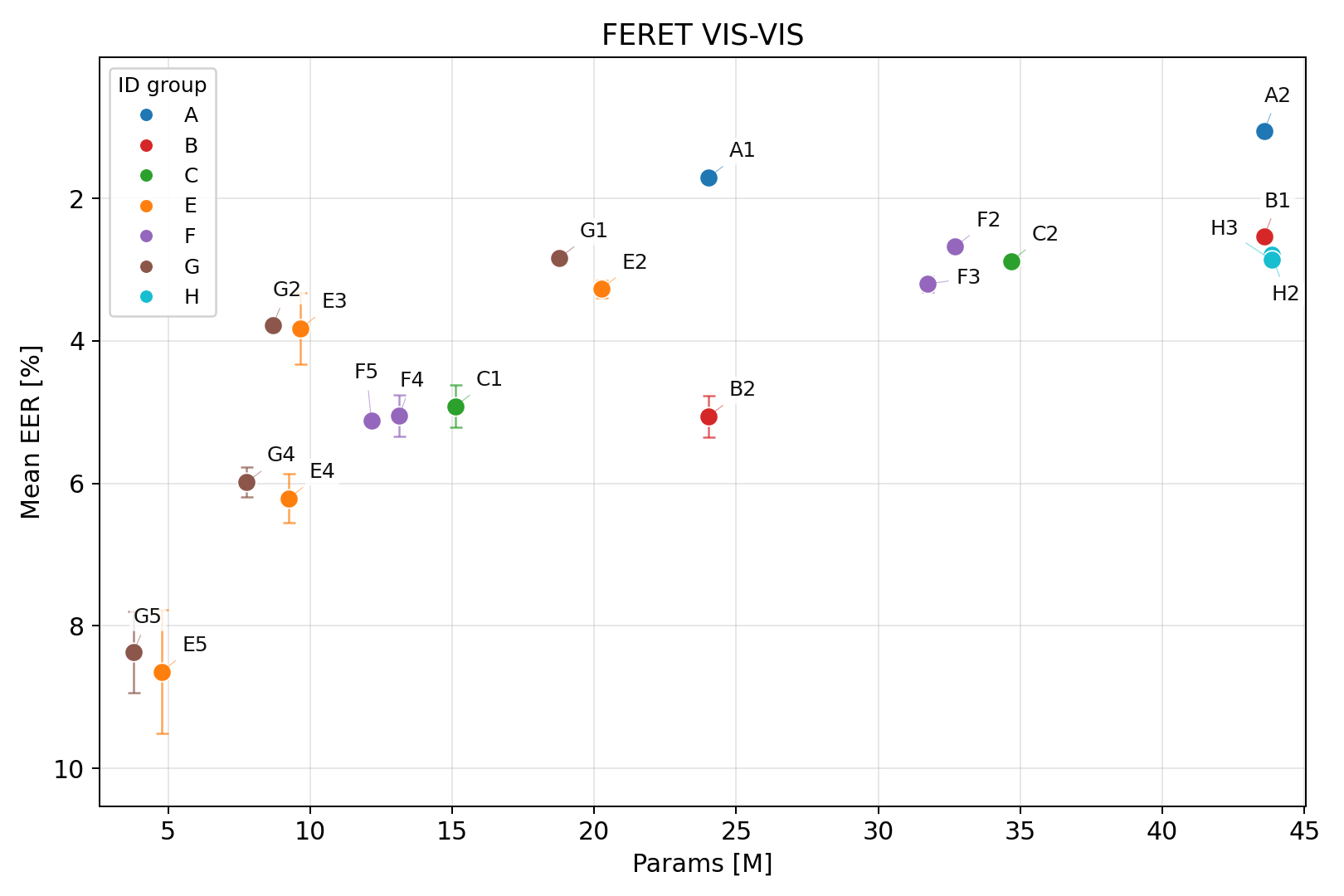}
\end{subfigure}
\caption{EER versus inference parameter count. Lower EER and fewer parameters are better. The plots separately represent capacity from image-level MAC cost, including the masked full-face baselines that retain full-face model parameters.}
\label{fig:eer_vs_params}
\end{figure*}

Figures~\ref{fig:fnmr1e3_vs_gmacs} and~\ref{fig:eer_vs_params} visualize complementary accuracy-efficiency trade-offs. Color points show means and error bars show sample standard deviations over successful runs; no standard deviation is available for a single successful run. Numerically failed runs are excluded from biometric averages.

Tables~\ref{tab:casia_visnir}-\ref{tab:feret} report the number of successful runs. The plots show that a substantial part of the computational reduction is obtained simply by moving from full-face $112\times112$ inputs to direct $80\times40$ periocular inputs. Further reductions in width and embedding size provide additional savings, but the biometric cost depends on the target protocol and operating point.

\subsection{Input Reduction: Masked Reuse versus Direct Periocular Training}
The input-region variants separate information content from recognition-stage efficiency. The full-face references \textit{A1} and \textit{A2} retain the original $112\times112$ recognition input. 

\textit{B1} and \textit{B2} evaluate a practical reuse baseline: the same full-face-trained model weights are evaluated with all pixels outside the periocular \textit{ROI} set to zero (masked). This evaluation is not a matched training study of masked $112\times112$ inputs versus direct $80\times40$ crops, because \textit{B1} and \textit{B2} reuse full-face-trained weights whereas \textit{C1} (IR-18) and \textit{C2} (IR-50) are trained directly on periocular crops. Instead, it asks whether simple full-face recognizer reuse remains competitive when only the eye region is visible.

Under this protocol, direct periocular training is usually preferable to masked full-face reuse. For \textit{IR-50}, \textit{C2} improves over \textit{B1} on \textit{CASIA VIS--NIR}, \textit{ND-NIVL VIS--NIR} and \textit{UFPR} while reducing recognition cost from $6.30$ to $1.64$ GMACs and increasing inference throughput from $8,107$ to $25,295$ images. 

The \textit{FERET VIS--VIS} is the main \textit{EER} exception, where \textit{B1} remains stronger; at FMR=$10^{-3}$, however, the FNMR values are nearly tied. For \textit{IR-18}, \textit{C1} improves \textit{EER} over \textit{B2} on all evaluated protocols and also lowers FNMR at FMR=$10^{-3}$ on \textit{CASIA}, \textit{ND-NIVL} and \textit{UFPR}, but not \textit{FERET} while reducing cost from $2.60$ to $0.68$ \textit{GMACs}.

These results suggest that direct periocular models are not only cheaper because they remove unused input area; they are also better adapted to the reduced visible region than full-face recognizers evaluated under zero-masked inputs. A likely explanation is train-test mismatch: the masked baselines are trained on complete aligned faces but evaluated with most non-periocular pixels suppressed, whereas the direct models observe compact periocular inputs during both training and evaluation.

\subsection{Depth, Width and Embedding Scaling}
As defined, depth scaling changes the number of residual units while preserving the input definition and loss. The \textit{IR-18} periocular model (\textit{C1}) is much faster than the \textit{IR-50} periocular baseline (\textit{C2}), but it loses accuracy on most protocols. Width scaling provides a smoother trade-off than switching directly from \textit{IR-50} to \textit{IR-18}. Because the width multiplier reduces channel counts throughout the stem and residual stages, it lowers both parameters and convolutional MACs substantially. The width \textit{0.75} model (\textit{E2}) reduces parameters by about 42\% relative to \textit{C2} and reduces MACs by about 44\%, while preserving competitive VIS--NIR performance. The width \textit{0.5} model (\textit{E3}) is considerably cheaper, but shows a larger degradation on \textit{UFPR} and \textit{FERET}.

Embedding-size reduction directly reduces template size and the parameters and \textit{MACs} of the final fully connected projection. The convolutional backbone remains unchanged, so the reduction in total inference \textit{MACs} is comparatively small. This is visible in \textit{F2} and \textit{F3}: reducing the embedding from $512$ to $256$ or $128$ dimensions approximately halves or quarters the template size, while throughput remains close to the baseline. The 256-dimensional model (\textit{F2}) is a useful trade-off; the \textit{128}-dimensional model (\textit{F3}) remains trainable but exhibits a clearer loss in accuracy.

\subsection{Combined Reductions and Failure Boundary}
The combined width and embedding reductions show where the model begins to reach the practical lower bound. The \textit{G1}, using width $0.75$ and a $256$-dimensional embedding, is one of the most attractive trade-offs: it reduces \textit{MACs} to $0.92$ \textit{GMACs} and the template to $1,024$ bytes while keeping \textit{EER} close to the stronger periocular models. The \textit{G2} further reduces width to $0.5$ and remains usable, but performance decreases.

More aggressive combinations with $128$-dimensional embeddings and reduced width did not converge reliably in our controlled training setup. We report these runs explicitly as diverged cases, marked \textit{DVG} in the tables, rather than omitting them. In these configurations, training became numerically unstable before a valid model state could be selected, so no verification metrics are reported. They are useful because they indicate that the reduced periocular input does not support arbitrary compression under the fixed \textit{AdaEye} training protocol.

Evaluation across seeds reveals numerical instability in less aggressively reduced configurations as well. Across the $18$ configurations evaluated using seeds $42$, $50$ and $51$, $45$ runs completed successfully and eight numerical failures were observed: \textit{F2} and \textit{F3} at seed $51$, \textit{F5} and \textit{G1} at seeds $50$ and $51$, \textit{G2} at seed $50$, and \textit{G4} at seed $51$. All three seeds in the \textit{A}, \textit{C}, \textit{E} and \textit{H} groups completed successfully. These findings distinguish verification performance after successful training from training reliability under the fixed protocol.

Failed runs contribute no verification scores. Consequently, \textit{F3}, \textit{G2} and \textit{G4} have two contributing runs, whereas \textit{F5} and \textit{G1} contribute only the result from seed $42$. For \textit{F2}, only seed $50$ contributes within the selected seed set.

\subsection{Final Spatial Representation}
The \textit{H2} and \textit{H3} evaluate whether the default late $3\times5$ spatial representation is too restrictive for $80\times40$ inputs. These models increase parameters and \textit{MACs} relative to the direct periocular baseline, but improve mean \textit{EER} and FNMR at FMR=$10^{-3}$ on \textit{CASIA} and \textit{ND-NIVL} relative to \textit{C2}, without a consistent improvement on \textit{UFPR}. This suggests that the spatial bottleneck is relevant for low-resolution periocular inputs: preserving more late-stage spatial information can improve recognition, but it moves the model away from the most efficient operating point.

\section{Discussion}
\label{sec:discussion}

The results distinguish three properties of a reduced recognition model: biometric performance after successful training, recognition-stage computational cost, and reliability of training under the fixed protocol. These properties need not improve together. Reducing input size or model capacity saves resources, but the resulting accuracy changes depend on the scaling axis, dataset and operating point.

\subsection{Scaling Effects and Protocol Dependence}
Moving from the full-face \textit{IR-50} reference \textit{A2} to the direct periocular baseline \textit{C2} reduces recognition cost from $6.30$ to $1.64$ GMACs/image, but entails substantial accuracy losses on the face-based protocols, especially at stringent FMRs. UFPR behaves differently: \textit{C2} improves mean EER from $3.083\%$ to $2.020\%$. Since \textit{UFPR} contains native periocular images, its full-size input is not a complete face; this result concerns adaptation to periocular content rather than a general advantage over full-face recognition.

The scaling axes affect different resources. Width reduction lowers both parameter count and convolutional cost, whereas smaller embeddings primarily reduce template storage and the training classification head. For example, reducing width from \textit{C2} to \textit{E2} lowers GMACs by approximately $44\%$. Mean EER changes from $1.358\%$ to $1.283\%$ on CASIA and from $0.697\%$ to $0.687\%$ on ND-NIVL, but increases from $2.020\%$ to $2.505\%$ on UFPR and from $2.880\%$ to $3.273\%$ on FERET. These differences do not establish statistical superiority, but show that reductions do not uniformly worsen every metric. Conversely, enlarging the final spatial representation in \textit{H2/H3} improves mean \textit{EER} and FNMR at FMR=$10^{-3}$ on \textit{CASIA} and ND-NIVL relative to \textit{C2}, without a consistent advantage on UFPR.

Training uses VIS data, yet the reduced models retain useful verification performance on both VIS--NIR protocols. The width-scaling results therefore do not suggest a uniformly greater penalty on the evaluated cross-spectral protocols. However, the VIS--VIS and VIS--NIR protocols use different datasets, so these comparisons do not isolate spectral mismatch from other dataset characteristics. Robustness also depends on the operating point: \textit{FERET-E3} has an \textit{EER} of $3.827\pm0.506\%$, but FNMR at FMR=$10^{-4}$ of $55.664\pm30.611\%$. Stability at \textit{EER} should therefore not be interpreted as stability throughout the low-\textit{FMR} region.

\subsection{Training Reliability and Practical Model Selection}
Training across multiple seeds shows that successful optimization is not guaranteed for every configuration. Numerical failures occur in the \textit{F} and \textit{G} groups, whereas all three seeds complete successfully in the \textit{A}, \textit{C}, \textit{E} and \textit{H} groups. Training instability is therefore not a monotonic consequence of lower parameter count alone. \textit{G1} illustrates this distinction: its successful seed-42 run combines useful verification performance with $0.92$ GMACs/image and a $1,024$-byte template, while seeds 50 and 51 fail numerically. Its favorable accuracy-cost trade-off must therefore be considered alongside its limited training reliability.

Model selection consequently depends on the required operating point and resources. The \textit{E2} combines reduced compute with comparable mean performance on the evaluated VIS--NIR protocols and successful training at all three seeds; F variants target template storage; \textit{H2/H3} trade additional computation for improvements on selected protocols. No configuration dominates across all these criteria. Means and standard deviations describe successful runs only and must be interpreted alongside the reported failures. Computational measurements concern recognition-stage cost and H100 throughput at batch size 512; they do not establish end-to-end latency or performance on mobile hardware.

\section{Conclusion}
\label{sec:conclusion}

AdaEye shows that periocular input and backbone scaling offer substantial recognition-stage resource savings, but their biometric consequences depend on the scaling axis, dataset and operating point. Direct periocular inputs reduce computation, width reduction provides further parameter and MAC savings, and embedding reduction mainly reduces template storage. Some reductions retain comparable mean VIS--NIR performance, whereas others incur substantial losses, particularly at stringent FMRs. VIS-only training supports cross-spectral verification on the evaluated protocols, without establishing that spectral mismatch is generally insensitive to compression.

Evaluation across seeds identifies training reliability as a separate consideration: configurations with favorable accuracy-cost trade-offs after successful training can still fail numerically under other seeds. The resulting contribution is a characterization of complementary operating points and their limitations under a common training protocol, rather than a universal winning model. Practical selection should account jointly for verification performance, recognition-stage resource requirements and observed training reliability.
\vspace{-0.2cm}

\section*{Acknowledgements}

This research work has been partially funded by the European Union’s Horizon research and innovation programs under grant CarMen (101168325), and by the German Federal Ministry of Education and Research and by the Hessian Ministry of Higher Education, Research, Science and the Arts within their joint support of the National Research Center for Applied Cybersecurity ATHENE.
\vspace{-0.2cm}

\bibliographystyle{IEEEtran}
\bibliography{references}
\vspace{-5pt}

\begin{IEEEbiography}[{\includegraphics[width=0.9in,height=1.25in,clip,keepaspectratio]{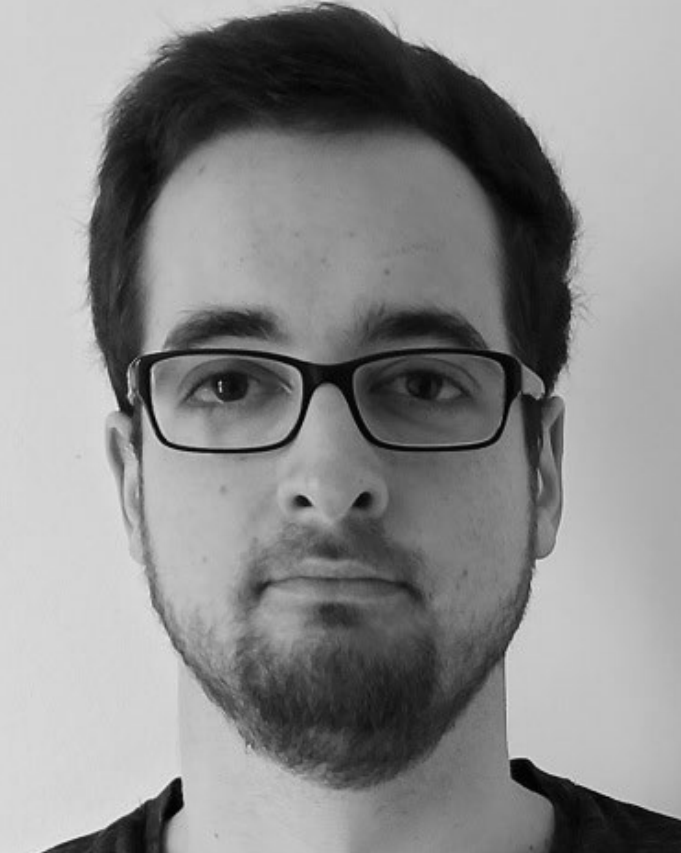}}]{Fabian Stockhardt}
is a member of the Darmstadt-based da/sec research group. He received the LL.B. and B.Sc. degrees from Hochschule Darmstadt. His main research interests include computer vision, biometrics, privacy and deep learning systems.
\end{IEEEbiography}
\vspace{-7pt}
\begin{IEEEbiography}[{\includegraphics[width=0.9in,height=1.25in,clip,keepaspectratio]{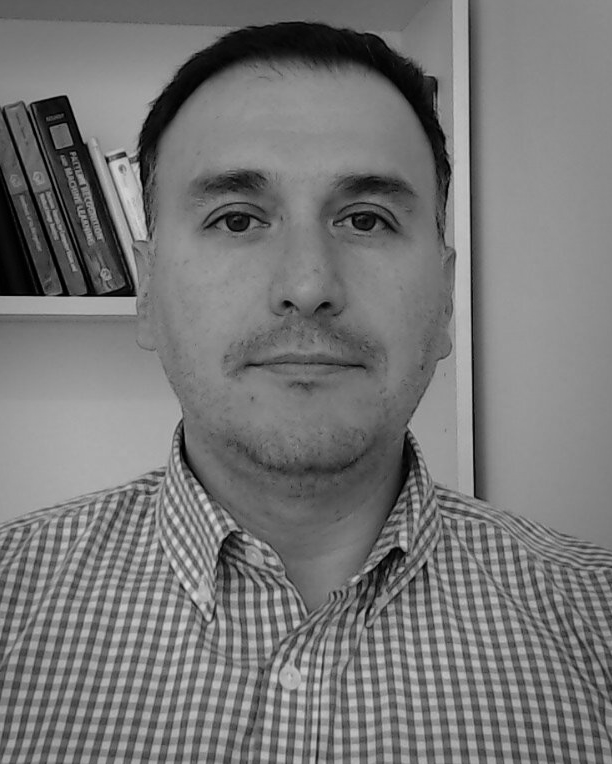}}]{Juan E. Tapia}
is a Senior Researcher at Hochschule Darmstadt (h-da) in Germany, where he leads EU-funded projects like EINSTEIN, iMARS, and CARMEN in biometrics and digital forensics. With over 15 years of experience in artificial intelligence, computer vision, and identity security, he is an entrepreneur and leading expert in Iris and Face Recognition, Presentation Attack Detection (PAD) for ID documents, Fitness-For-Duty evaluation (FFD), and generative AI for biometrics. He holds a PhD in Electrical Engineering from the Universidad de Chile (2016), an MSc (2012), and a BSc in Electronics Engineering from Universidad Mayor (2004). He was awarded the Best PhD Thesis Award in 2016. Professor at several Universities in Chile in the computer science department. Dr. Tapia is an active reviewer for top-tier journals and conferences and serves as an Associate Editor for the IET Image Processing Journal. He is also a member of the German DIN on behalf of the ISO/IEC SC-37 on biometrics. His research focuses on foundation models (ViT, GANs, diffusion models), deep learning, and pattern recognition applied to biometrics, morphing, and Presentation Attack Detection on iris, faces, and ID cards.
\end{IEEEbiography}
\vspace{-7pt}

\begin{IEEEbiography}[{\includegraphics[width=0.9in,height=1.25in,clip,keepaspectratio]{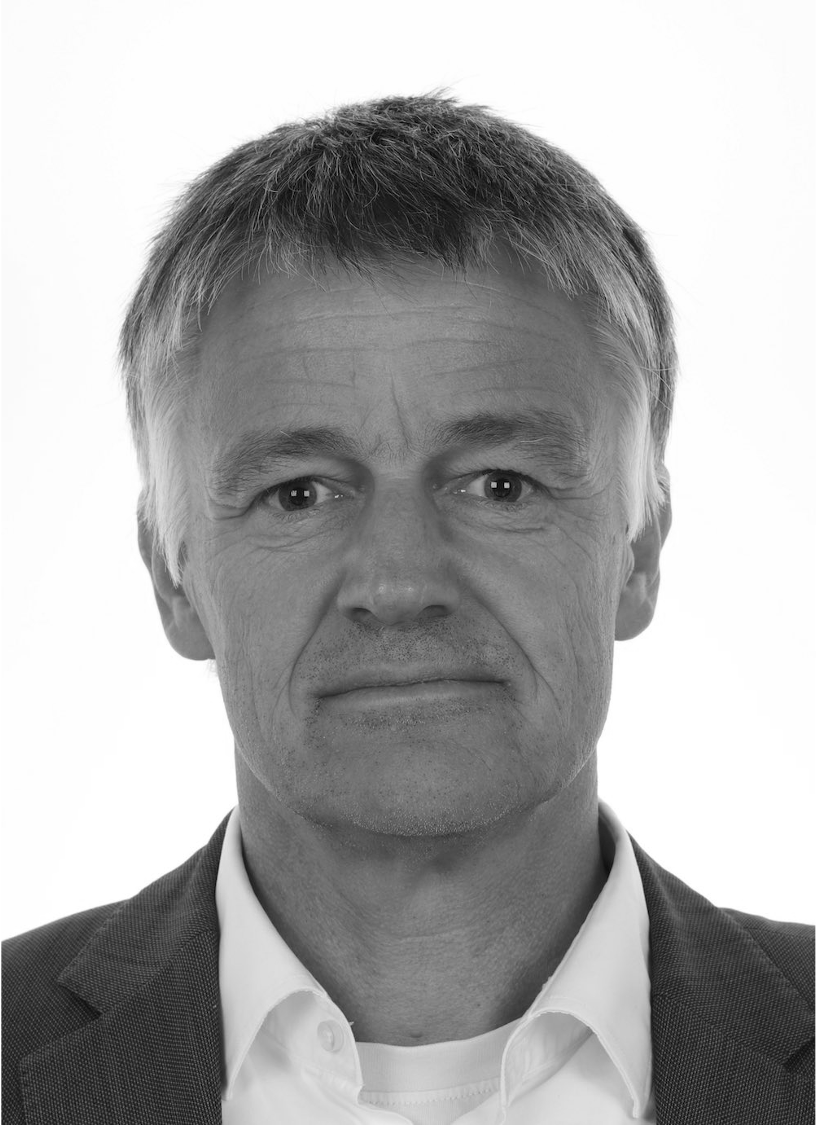}}]{Christoph Busch}
is a member of the Department of Information Security and Communication Technology (IIK) at the Norwegian University of Science and Technology (NTNU), Norway. He holds a joint appointment with the computer science faculty at Hochschule Darmstadt (HDA), Germany. On behalf of the German BSI, he has been the coordinator for the project series BioIS, BioFace, BioFinger, BioKeyS Pilot-DB, KBEinweg and NFIQ2.0. In the European research program, he was the initiator of the Integrated Project 3D-Face, FIDELITY and iMARS. Further, he was/is a partner in the projects TURBINE, BEST Network, ORIGINS, INGRESS, PIDaaS, SOTAMD, RESPECT and TReSPAsS. He is also the principal investigator at the German National Research Centre for Applied Cybersecurity (ATHENE). Moreover, Christoph Busch is a co-founder and board member of the European Association for Biometrics (www.eab.org), which was established in 2011 and comprises more than 200 institutional members. Christoph co-authored more than 700 technical papers and has been a speaker at international conferences. Christoph Busch is a member of the German Gesellschaft fur Informatik (GI) and the International Association of Pattern Recognition (IAPR), as well as an IEEE Fellow.
\end{IEEEbiography}

\vfill

\end{document}